\pdfoutput=1

\documentclass[11pt]{article}

\usepackage[preprint]{acl}

\usepackage{times}
\usepackage{latexsym}
\usepackage{booktabs}
\usepackage{array}
\usepackage[T1]{fontenc}
\usepackage{float}
\usepackage{booktabs}

\usepackage[utf8]{inputenc}
\usepackage{multirow}
\usepackage{multicol}
\usepackage{microtype}
\usepackage{inconsolata}

\usepackage{graphicx}
\usepackage{float}
\usepackage{amsmath}
\usepackage{subcaption}
\usepackage{xcolor} 
\usepackage[skins,breakable]{tcolorbox}
\usepackage{listings} 
\tcbset{
  remarkbox/.style={
    enhanced,
    breakable,
    colback=blue!5,
    colframe=blue!70!black,
    coltitle=white,
    fonttitle=\bfseries,
    title=Observation \#1,
    boxrule=0.8pt,
    arc=3pt,
    outer arc=3pt,
    left=5pt,
    right=5pt,
    top=4pt,
    bottom=4pt
  }
}

\title{Seeing is Believing, but How Much? A Comprehensive Analysis of Verbalized Calibration in Vision-Language Models}

\author{\textbf{Weihao Xuan}$^{1,2*}$, \textbf{Qingcheng Zeng}$^{3}$\thanks{Both authors contributed equally.}, \textbf{Heli Qi}$^{4}$, \textbf{Junjue Wang}$^{1}$, \textbf{Naoto Yokoya}$^{1,2}$\thanks{Corresponding author.} \\
\\
$^{1}$The University of Tokyo, $^{2}$RIKEN AIP, $^{3}$Northwestern University, $^{4}$Waseda University \\
}

\begin{document}
\maketitle
\begin{abstract}
Uncertainty quantification is essential for assessing the reliability and trustworthiness of modern AI systems. Among existing approaches, verbalized uncertainty—where models express their confidence through natural language—has emerged as a lightweight and interpretable solution in large language models (LLMs). However, its effectiveness in vision-language models (VLMs) remains insufficiently studied. In this work, we conduct a comprehensive evaluation of verbalized confidence in VLMs, spanning three model categories, four task domains, and three evaluation scenarios. Our results show that current VLMs often display notable miscalibration across diverse tasks and settings. Notably, visual reasoning models (i.e., \textit{thinking with images}) consistently exhibit better calibration, suggesting that modality-specific reasoning is critical for reliable uncertainty estimation. To further address calibration challenges, we introduce \textsc{Visual Confidence-Aware Prompting}, a two-stage prompting strategy that improves confidence alignment in multimodal settings. Overall, our study highlights the inherent miscalibration in VLMs across modalities. More broadly, our findings underscore the fundamental importance of modality alignment and model faithfulness in advancing reliable multimodal systems.
\end{abstract}

\section{Introduction}
Recent advances in large language models (LLMs) and vision-language models (VLMs) have led to significant progress across a broad spectrum of capabilities, including reasoning \cite{openai2024openaio1card, openai2025competitiveprogramminglargereasoning}, instruction following~\cite{zhou2023instructionfollowingevaluationlargelanguage, grattafiori2024llama3herdmodels}, and visual understanding~\cite{liu2023visual, li2024llavaov, padlewski2024vibeevalhardevaluationsuite, agrawal2024pixtral12b, bai2025qwen25vltechnicalreport}. However, as these models are increasingly deployed in real-world and high-stakes applications, evaluating their trustworthiness has become as essential as measuring their task performance. A fundamental aspect of this assessment is calibration - ensuring that a model's expressed confidence aligns with its actual accuracy in real-world scenarios. With the rise of closed-source models that only support text-based interactions, the ability to express uncertainty, similar to human communication verbally, has become particularly crucial for practical applications~\cite{xiong2024llmsexpressuncertaintyempirical}.

Although quite a few studies have explored eliciting more accurate confidence estimations from LLMs~\cite{lin2022teachingmodelsexpressuncertainty, tian-etal-2023-just, xu-etal-2024-sayself, hager2025uncertaintydistillationteachinglanguage}, how these strategies adapt to VLMs remains an open question. Unlike text-only models, VLMs process and integrate information across multiple modalities, introducing new dimensions of complexity in how confidence is expressed and calibrated. This multimodal nature raises three critical challenges: 1) How accurately can VLMs verbalize their confidence? 2) How do instructions embedded within images affect calibration? 3) Does verbalized confidence remain consistent when processing the same information presented in different modalities?

These questions highlight a gap in current research and underscore the need for a deeper investigation into modality-sensitive uncertainty estimation in VLMs.
Our contributions are threefold: 1) We present the first comprehensive evaluation of verbalized confidence in a diverse set of commercial and open-source VLMs, leveraging widely adopted large-scale multimodal datasets. Our analysis focuses on three evaluation scenarios, with particular focus on the embedded instruction setting and the semantically aligned setting, to investigate how calibration behaviors vary across input modalities. 2) Our results indicate that most VLMs continue to exhibit miscalibration in their verbalized uncertainty. However, visual reasoning models, those capable of \textit{thinking with images}, show notable improvements in calibration across benchmarks and modalities. These findings point to the promise of visual reasoning-oriented enhancement for verbalized uncertainty estimation. 3) To further improve calibration, we introduce \textsc{visual confidence-aware prompting}, a two-stage strategy that elicits visual-specific confidence before aggregating the final output. Compared to strong baselines such as Top-K prompting and self-reflection, our method yields significant gains in calibration quality. 

In summary, this work comprehensively evaluates the verbalized calibration in current VLMs, reveals persistent miscalibration issues across modalities, and offers a promising direction for enhancing verbalized uncertainty.

\begin{figure*}[ht!]
    \centering
    \includegraphics[width=\textwidth]{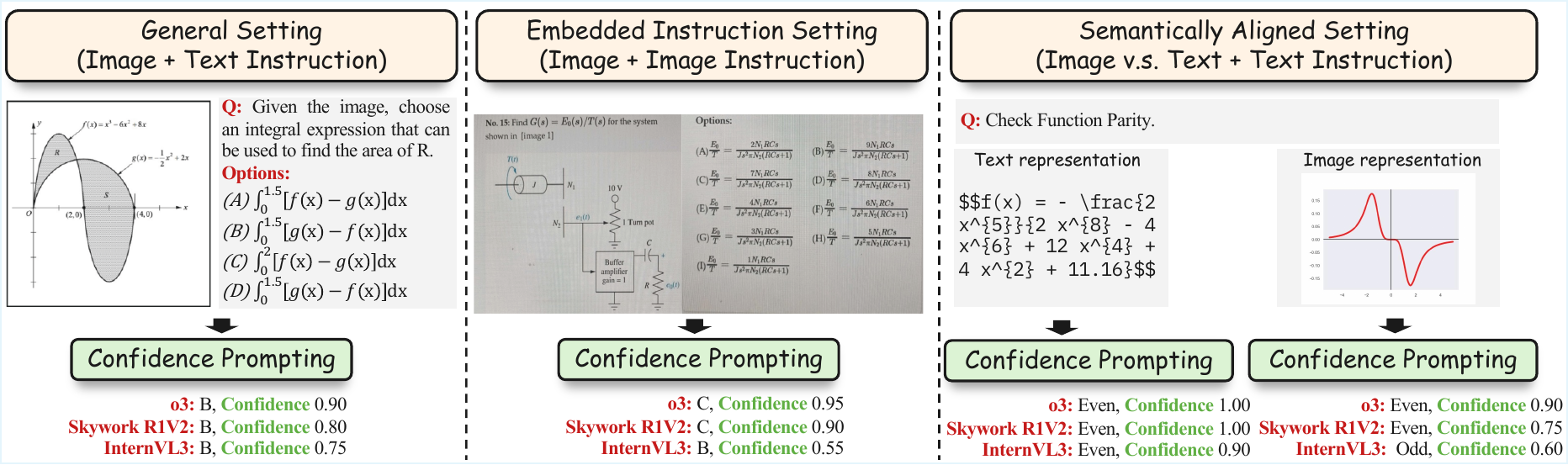}
    \caption{The illustration of our three types of evaluations: general, embedded instruction, and semantically aligned evaluation. These configurations test VLMs' calibration across different input modalities and instruction formats.}
    \label{fig:illu}
\end{figure*}

\section{Related Works}
\subsection{Uncertainty Quantification and Calibration in L(V)LMs}
Quantifying and calibrating uncertainty is a key area of research for improving the reliability of LLMs~\cite{fadeeva-etal-2023-lm, 10.1162/tacl_a_00737}. A variety of techniques have been explored to estimate uncertainty in L(V)LMs, including sampling-based approaches~\cite{kuhn2023semanticuncertaintylinguisticinvariances, nikitin2024kernellanguageentropyfinegrained}, which approximate uncertainty by drawing from predictive distributions; information-theoretic methods~\cite{fadeeva-etal-2024-fact, chen-etal-2025-unveiling}, which employ measures such as entropy or mutual information; and reflexive approaches~\cite{tian-etal-2023-just, xiong2024llmsexpressuncertaintyempirical}, which investigate how models articulate their own uncertainty through natural language.

This work focuses on the reflexive category, specifically examining how VLMs verbalize uncertainty. Verbalized uncertainty offers practical advantages: it eliminates the need for computational overhead associated with sampling or post-hoc calibration, while providing uncertainty assessments that are easily understood by general users~\cite{hager2025uncertaintydistillationteachinglanguage}. Prior studies have identified consistent patterns of~\textbf{overconfidence} in verbalized uncertainty produced by instruction-tuned models~\cite{xiong2024llmsexpressuncertaintyempirical, yang2024verbalizedconfidencescoresllms}. Others have observed that reinforcement learning (RL)-based reasoning can improve calibration across domains, suggesting that RL may induce more reflective and self-aware model outputs~\cite{zeng2025reasoningmodelsbetterverbalized}. While initial analyses of verbalized uncertainty in VLMs have been conducted, primarily on small-scale or object-level datasets~\cite{groot-valdenegro-toro-2024-overconfidence, borszukovszki-etal-2025-know, zhao2025objectlevelverbalizedconfidencecalibration}, a broader, systematic evaluation across tasks, domains, and model families remains lacking. Our work addresses this gap through a holistic analysis of how VLMs express and calibrate their uncertainty across diverse settings.

\subsection{Modality Misalignment in Multimodal LLMs}
In multimodal LLMs, an ideal expectation is that the models maintain consistent performance when equivalent information is presented across different modalities. However, a growing body of research in vision-language~\cite{10.1162/tacl_a_00698, mistretta2025crossgapexposingintramodal, shu2025largevisionlanguagemodelalignment} and audio-language settings \cite{chen2024voicebenchbenchmarkingllmbasedvoice} has highlighted significant modality misalignment in many current multimodal models. These findings demonstrate systematic failures in cross-modal information integration and generalization. In response, several benchmarks have been proposed to quantify and analyze these performance gaps~\cite{fu2024isobenchbenchmarkingmultimodalfoundation}. In this work, we extend the analysis to examine how calibration, particularly through verbalized uncertainty, behaves across text and image modalities in VLMs, providing novel insights into both alignment and confidence consistency.

\section{Experimental Setup}
\subsection{Evaluation Configurations}

To comprehensively analyze verbalized confidence in VLMs, we design three complementary evaluation configurations, each targeting distinct dimensions of model behavior related to modality, instruction presentation, and calibration robustness across different modalities (\autoref{fig:illu}). 
These settings allow us to assess not only the overall calibration performance but also how it varies with changes in input format and reasoning demands.

\subsubsection{General Evaluation}
In the general evaluation setting, we assess verbalized confidence when VLMs are prompted via textual instructions and required to reason over visual inputs. This configuration reflects a common usage scenario in which users provide tasks through natural language while the model processes accompanying images or video. We evaluate model calibration across four major task types:

\begin{itemize}
    \item \textbf{Image Understanding and Reasoning}: We adopt the MMMU-Pro benchmark \cite{yue2024mmmuprorobustmultidisciplinemultimodal} to measure calibration on complex, multidisciplinary image understanding and reasoning problems. This benchmark encompasses a wide array of domains and task formats, providing a robust benchmark for verbalized uncertainty.
    \item \textbf{Video Understanding and Reasoning}: To investigate video understanding and the corresponding calibration over dynamic visual content, we evaluate models using VideoMMMU \cite{hu2025videommmuevaluatingknowledgeacquisition}, covering perception, comprehension, and adaptation tasks. For fair comparison, we uniformly sample 32 frames per video as model input across all evaluations.
    \item \textbf{Factuality}: We adopt the Visual SimpleQA benchmark \cite{wang2025visualsimpleqabenchmarkdecoupledevaluation} to assess models' ability to judge factual correctness from visual information, offering a direct test of basic visual grounding and confidence estimation.
    \item \textbf{Math Reasoning}: To evaluate calibration under visual mathematical reasoning, we employ MathVista~\cite{lu2024mathvista} and MathVision~\cite{wang2024measuring}. These datasets involve interpreting diagrams and solving quantitative problems. We use the \textit{testmini} splits from both datasets to balance computational efficiency with sufficient task diversity.
\end{itemize}

\subsubsection{Embedded Instruction Evaluation}
\label{embedded}

A growing line of research has explored whether VLMs can accurately interpret instructions when they are embedded within visual inputs, rather than provided through the standard text modality~\cite{li2024textimagesmultimodallarge}. While most prior work focuses on task performance, less is known about how such modality shifts affect models’ verbalized calibration. To address this, we adopt the vision-only configuration from MMMU-Pro \cite{yue2024mmmuprorobustmultidisciplinemultimodal}, in which entire questions are embedded within images and directly presented to the model. By comparing calibration performance against the general setting, where question bodies are given via text, we assess whether visually embedded instructions introduce additional difficulty for VLMs in producing reliable confidence estimates.

\vspace{-2mm}
\subsubsection{Semantically Aligned Modalities Evaluation}
\label{semanticaligned}

Building on the previous setting, we take a further step toward disentangling modality effects by evaluating VLMs on inputs that are semantically equivalent but presented in different modalities. For example, a mathematical function may appear either as a visual diagram or a textual equation (see \autoref{fig:illu}). This setup allows us to isolate calibration and reasoning behavior when the content remains constant, but the modality changes. To this end, we use the IsoBench benchmark \cite{fu2024isobenchbenchmarkingmultimodalfoundation}, which spans four domains (mathematics, games, science, and algorithms) and is explicitly designed for testing modality alignment. Unlike the embedded instruction setting, which focuses on instruction modality, this scenario enables a more nuanced analysis of modality-specific reasoning gaps, revealing whether VLMs process and calibrate equivalent information differently depending on its format.

\subsection{Models}
We evaluate a broad selection of state-of-the-art VLMs, spanning both commercial and open-source models. To better understand how training objectives and reasoning styles affect verbalized calibration, we categorize the models into three groups based on their alignment strategies and dominant reasoning modalities:
\begin{itemize}
\item[1.] \textbf{General Instruction-Tuned Models:}
These models are optimized for following human instructions across a wide range of multimodal tasks. They are typically trained with supervised fine-tuning and enhanced with preference alignment. Representative models in this category include OpenAI GPT-4.1 and GPT-4o \cite{openai2024gpt4ocard}, Qwen-VL series (Qwen-VL-2/2.5 in both 7B and 72B scales) \cite{bai2025qwen25vltechnicalreport}, InternVL3-78B \cite{zhu2025internvl3exploringadvancedtraining}, and Kimi-VL-A3B-Instruct \cite{kimiteam2025kimivltechnicalreport}.
\item[2.] \textbf{Text-Centric Reasoning Models:}  
This group includes models that primarily reason over textual representations, often enhanced via reinforcement learning or instruction tuning with an emphasis on chain-of-thought or self-reflective reasoning. These models typically generate intermediate reasoning steps in text before producing answers. Included here are OpenAI o1 \cite{openai2024openaio1card}, Kimi-VL-A3B-Thinking \cite{kimiteam2025kimivltechnicalreport}, Skywork-R1V-38B \cite{peng2025skyworkr1vpioneeringmultimodal}, and Skywork-R1V2-38B \cite{chris2025skyworkr1v2multimodalhybrid}.
\item[3.] \textbf{Vision-Centric Reasoning Models:}  
These models are explicitly designed to perform visual chain-of-thought reasoning, where multimodal inputs, particularly images, are not only used for grounding but also as part of the model’s internal reasoning process. OpenAI o3 and o4-mini \cite{o3} fall into this category, as they are trained to natively integrate visual elements into multi-step reasoning workflows.
\end{itemize}
This categorization allows us to analyze how verbalized calibration varies depending on whether reasoning is primarily text-driven, vision-driven, or instruction-oriented, offering a clearer lens into the strengths and limitations of different model families.

\subsection{Evaluation Metrics}
We examine the use of confidence scores from LLMs: calibration and failure prediction. Calibration evaluates whether the model's predicted confidence reflects its true likelihood of being correct. For example, when a model assigns 90\% confidence to its answers, it should be accurate approximately 90\% of the time. This alignment is especially important for applications that rely on reliable uncertainty estimates, such as safety-critical systems or human–AI collaboration.

To assess calibration, we use the Expected Calibration Error (ECE), which captures the average difference between predicted confidence and observed accuracy across $M$ bins:
\begin{equation}
    ECE = \sum_{m=1}^{M} \frac{|B_m|}{n} \left| \text{acc}(B_m) - \text{avgConf}(B_m) \right|
\end{equation}

Here, $n$ is the total number of examples and $B_m$ denotes the set of samples in the $m$-th confidence bin. We use $M = 10$ bins in all cases and compute ECE only over attempted questions when evaluating on factuality datasets. 

\section{Results}
\subsection{General Evaluation}
Our evaluation results of the general setting are presented in \autoref{tab:comprehensive_benchmarks}. Overall, most models exhibit moderate calibration performance, with a consistent tendency toward miscalibration.

\begin{table*}[htbp]
\centering
\renewcommand{\arraystretch}{0.8}
\resizebox{\textwidth}{!}{%
\small{
\begin{tabular}{llcccccc}
\toprule
\multirow{2}{*}{Metric} & \multirow{2}{*}{Model} & MMMU-Pro & VideoMMMU & Visual SimpleQA & \multirow{2}{*}{MathVista} & \multirow{2}{*}{MathVision} \\
& & (Standard, 10) & (P/C/A) & (Multimodal) & & \\
\midrule
\multirow{15}{*}{ACC $\uparrow$}
       & \multicolumn{6}{l}{\textit{Visual Reasoning Models}} \\
\cmidrule{2-7}
       & o3                    & 73.7  & 74.6/71.2/41.7 & 73.6 & 50.0 & 56.0 \\
       & o4-mini               & 68.7  & 72.8/67.5/39.2 & 66.5 & 48.5 & 52.4 \\
\cmidrule{2-7}
       & \multicolumn{6}{l}{\textit{Text Reasoning Models}} \\
\cmidrule{2-7}
       & o1                    & 70.4  & 72.7/66.2/40.3 & 70.6 & 46.7 & 51.0 \\
       & Skywork-R1V2 38B      & 55.2  & 59.9/58.6/40.7 & 45.5 & 44.7 & 39.6 \\
       & Kimi-VL-A3B-Thinking  & 45.2  & 62.0/55.1/32.6 & 39.3 & 46.7 & 30.5 \\
\cmidrule{2-7}
       & \multicolumn{6}{l}{\textit{Instruct Models}} \\
\cmidrule{2-7}
       & GPT4.1                & 65.0  & 74.9/62.3/40.8 & 67.1 & 47.9 & 43.0 \\
       & Qwen2.5-VL 7B         & 38.7  & 66.9/52.3/31.3 & 32.2 & 45.5 & 24.9 \\
       & Qwen2.5-VL 72B        & 53.8  & 80.0/69.7/44.3 & 49.6 & 49.9 & 40.8 \\
       & InternVL3 78B         & 55.1  & 66.7/54.7/35.8 & 44.0 & 47.1 & 34.2 \\
       & Kimi-VL-A3B-Instruct  & 38.7  & 72.3/41.7/30.7 & 35.3 & 42.6 & 28.3 \\
\midrule
\multirow{15}{*}{ECE $\downarrow$}
       & \multicolumn{6}{l}{\textit{Visual Reasoning Models}} \\
\cmidrule{2-7}
       & o3                    & 0.047 & 0.073/0.051/0.092 & 0.085 & 0.242 & 0.111 \\
       & o4-mini               & 0.174 & 0.125/0.172/0.293 & 0.069 & 0.388 & 0.327 \\
\cmidrule{2-7}
       & \multicolumn{6}{l}{\textit{Text Reasoning Models}} \\
\cmidrule{2-7}
       & o1                    & 0.245 & 0.204/0.271/0.470 & 0.145 & 0.474 & 0.447 \\
       & Skywork-R1V2 38B      & 0.312 & 0.233/0.271/0.403 & 0.365 & 0.438 & 0.427 \\
       & Kimi-VL-A3B-Thinking  & 0.433 & 0.343/0.371/0.553 & 0.476 & 0.475 & 0.598 \\
\cmidrule{2-7}
       & \multicolumn{6}{l}{\textit{Instruct Models}} \\
\cmidrule{2-7}
       & GPT4.1                & 0.321 & 0.225/0.342/0.549 & 0.275 & 0.485 & 0.535 \\
       & Qwen2.5-VL 7B         & 0.496 & 0.235/0.347/0.548 & 0.252 & 0.418 & 0.660 \\
       & Qwen2.5-VL 72B        & 0.392 & 0.161/0.248/0.443 & 0.371 & 0.466 & 0.580 \\
       & InternVL3 78B         & 0.387 & 0.278/0.379/0.550 & 0.402 & 0.480 & 0.617 \\
       & Kimi-VL-A3B-Instruct  & 0.492 & 0.203/0.488/0.571 & 0.421 & 0.482 & 0.612 \\
\bottomrule
\end{tabular}%
}}
\caption{Performance metrics across different datasets and models with CoT prompting. All accuracy (Acc) values are in percentage.}
\label{tab:comprehensive_benchmarks}
\vspace{-5mm}
\end{table*}

\begin{figure}[h]
    \centering
    \includegraphics[width=0.95\linewidth]{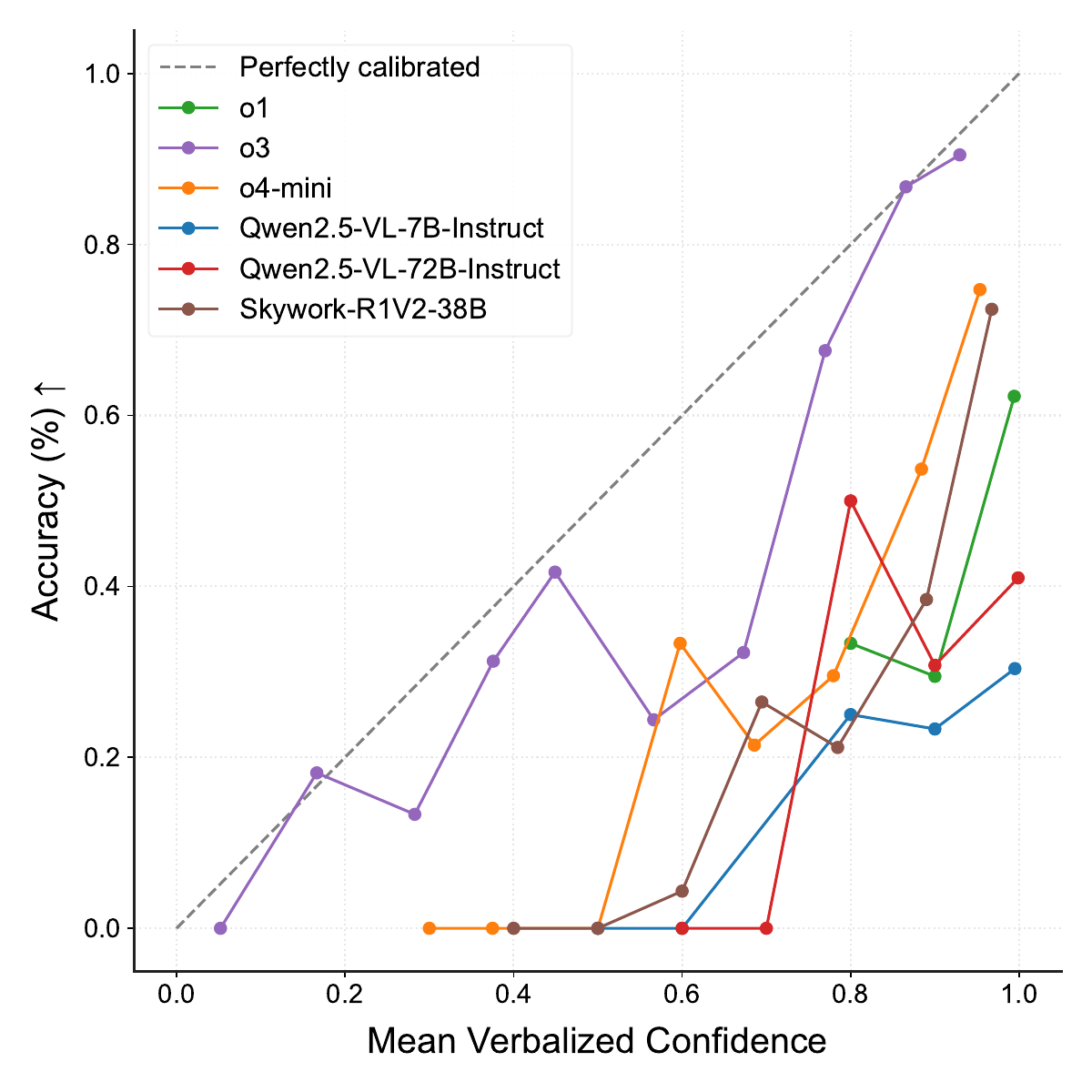}
    \caption{Calibration curve on testmini set of MathVision.}
    \label{fig:calibration_curve}
    \vspace{-4mm}
\end{figure}

Across different datasets, most instruct and text reasoning models exhibit ECE scores exceeding 0.25 in the majority of settings, indicating a clear tendency toward miscalibration. Moreover, the calibration curves presented in \autoref{fig:calibration_curve} show that these models are systematically overconfident across a wide range of confidence bins. In contrast, the \textit{o3} model consistently demonstrates strong calibration across all evaluated tasks, suggesting that it is less susceptible to the overconfidence patterns commonly observed in other VLMs.

When comparing model categories, visual reasoning models consistently demonstrate better calibration than both text-centric reasoning models and instruction-tuned models. Specifically, \textit{o3} and \textit{o4-mini}, which are optimized for visual reasoning, produce lower ECE scores across benchmarks, reflecting more reliable confidence estimates. In contrast, instruction-following and text-based reasoning models tend to exhibit higher ECE values, indicating less accurate self-assessment. This difference is especially notable in the two mathematical benchmarks, where all VLMs achieve similar levels of accuracy, yet visual reasoning models are significantly better calibrated.

Additionally, when comparing instruction-tuned models with similarly scaled text reasoning models (e.g., GPT-4.1 vs. o1, Kimi-VL-Instruct vs. Kimi-VL-Thinking), we observe that reasoning-oriented models tend to show slightly improved calibration. This suggests that reinforcement learning and reasoning-focused training can enhance a model's ability to assess its own uncertainty, particularly within the same modality, though modest benefits may also emerge from enhancements within the text modality. Taken together, these findings highlight the complementary roles of modality-specific and reasoning-specific training in building VLMs with more trustworthy and well-calibrated verbalized confidence.

\begin{table*}[htbp]
\centering
\renewcommand{\arraystretch}{0.8}
\resizebox{\textwidth}{!}{%
\small{
\begin{tabular}{ll*{5}{c}}
\toprule
Metric & Model & \multicolumn{5}{c}{IsoBench} \\
\cmidrule(lr){3-7}
& & Mathematics & Games & Science & Algorithms & All \\
\midrule
\multirow{16}{*}{ECE $\downarrow$}
       & \multicolumn{6}{l}{\textit{Visual Reasoning Models}} \\
\cmidrule{2-7}
       & o3 & 0.088/0.075 & 0.162/0.106 & 0.094/0.113 & 0.034/0.085 & 0.037/0.081 \\
       & o4-mini & 0.054/0.025 & 0.283/0.309 & 0.022/0.026 & 0.083/0.034 & 0.110/0.058 \\
\cmidrule{2-7}
       & \multicolumn{6}{l}{\textit{Text Reasoning Models}} \\
\cmidrule{2-7}
       & o1 & 0.187/0.007 & 0.522/0.425 & 0.080/0.033 & 0.223/0.022 & 0.265/0.109 \\
       & Skywork-R1V2-38B & 0.304/0.008 & 0.529/0.368 & 0.094/0.025 & 0.420/0.151 & 0.364/0.120 \\
       & Kimi-VL-A3B-Thinking & 0.376/0.077 & 0.708/0.611 & 0.117/0.044 & 0.554/0.540 & 0.462/0.284 \\
\cmidrule{2-7}
       & \multicolumn{6}{l}{\textit{Instruct Models}} \\
\cmidrule{2-7}
       & GPT4.1 & 0.111/0.007 & 0.520/0.460 & 0.081/0.024 & 0.165/0.079 & 0.216/0.131 \\
       & Qwen2.5-VL 7B & 0.377/0.069 & 0.695/0.629 & 0.155/0.059 & 0.526/0.469 & 0.467/0.284 \\
       & Qwen2.5-VL 72B & 0.353/0.008 & 0.724/0.534 & 0.146/0.039 & 0.376/0.320 & 0.431/0.200 \\
       & InternVL3 78B & 0.339/0.007 & 0.684/0.539 & 0.110/0.042 & 0.386/0.307 & 0.412/0.198 \\
       & Kimi-VL-A3B-Instruct & 0.327/0.257 & 0.651/0.604 & 0.068/0.032 & 0.471/0.503 & 0.411/0.372 \\
\bottomrule
\end{tabular}%
}}
\caption{IsoBench results across different categories and models with CoT prompting. Image/text modality results are shown with slash (/). For Mathematics, the text modality shows LaTeX format results; for Games, the text modality shows PGN format results.}
\label{tab:grouped_isobench}
\end{table*}

\subsection{Embedded Instruction Evaluation}
\begin{figure}[h]
    \centering
    \includegraphics[width=\columnwidth]{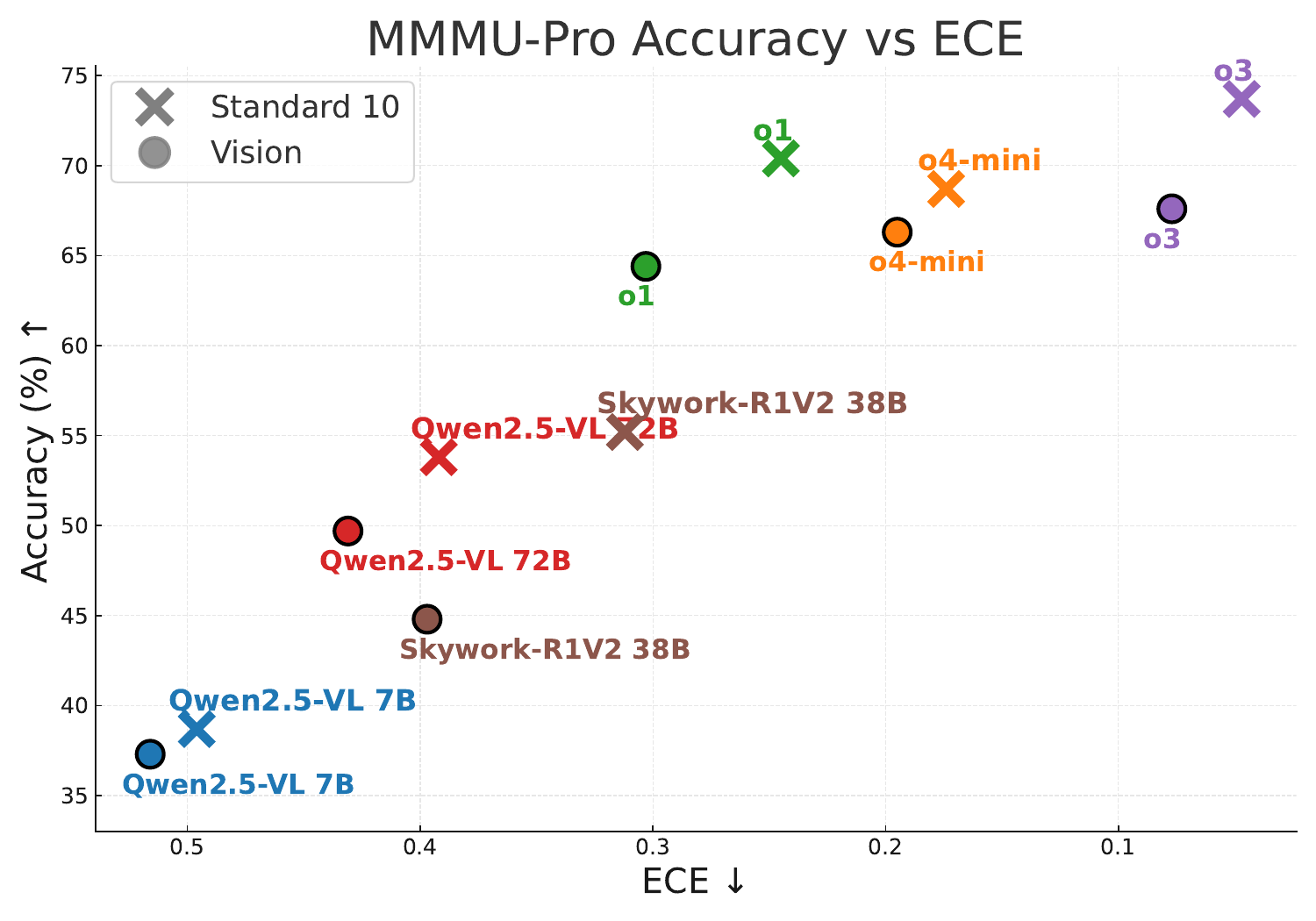}
    \caption{Model performance comparison across and calibration. The upper right indicates better overall performance.}
    \label{fig:mmmupro}
    \vspace{-5mm}
\end{figure}

Here, we present results from the embedded instruction setting, where question bodies are provided exclusively through visual inputs. The results are visualized in \autoref{fig:mmmupro}, where data points from the vision-based setting are marked with circles, and those from the general (text-instruction) setting are marked with crosses, allowing for direct visual comparison.

Overall, our findings indicate that most evaluated VLMs continue to face challenges in calibration when processing visually embedded instructions. The instruction-tuned models still exhibit ECE scores above 0.4, reflecting a notable degree of miscalibration. In contrast, reasoning-oriented models demonstrate better calibration performance, with visual reasoning models achieving the most reliable confidence estimates. These results further support our earlier observation that both modality-specific and reasoning-specific training play a critical role in improving verbalized uncertainty in VLMs.

When comparing performance under visual instructions to the general evaluation setting, we observe a clear drop in both accuracy and calibration. Even the strongest models, such as \textit{o3} and \textit{o4-mini}, exhibit noticeable degradation when instructions are presented visually rather than textually. These findings point to a persistent misalignment between vision and language modalities in current VLMs, indicating that interpreting visual instructions remains a core challenge in multimodal understanding. Addressing this modality gap is crucial for developing more robust and trustworthy VLMs capable of producing consistent and calibrated confidence estimates across diverse input formats.

\subsection{Semantically Aligned Evaluation}
In our final evaluation setting, we assess calibration performance when VLMs are given identical textual instructions paired with semantically equivalent inputs presented in either the textual or visual modality for reasoning. The corresponding ECE results are reported in \autoref{tab:grouped_isobench}.

Consistent with our previous findings, we observe a substantial calibration gap between modalities in most models. Although the underlying content remains identical, VLMs tend to produce less calibrated responses when reasoning over visual inputs compared to text. This discrepancy is especially evident in the \textit{mathematics} domain, where text-based inputs are relatively easy for VLMs to solve, but performance degrades noticeably with visual input. In these cases, models often assign high confidence regardless of their actual visual reasoning capabilities, leading to significantly higher ECE scores. The only notable exceptions are visual reasoning models, which consistently exhibit smaller calibration gaps across modalities when compared to instruction-tuned and text-centric reasoning models. We also found a particularly difficult task, puzzle in the game category, although state-of-the-art models all show accuracy lower than 10\%, only \textit{o3} shows moderate level calibration, suggesting the great potential of vision-based reasoning in reducing miscalibration.

Taken together, our experiments across the three evaluation settings provide a comprehensive picture of how current VLMs handle verbalized uncertainty under varying input structures. In the general setting, most models show moderate calibration, though miscalibration remains widespread. In the embedded instruction setting, calibration performance declines noticeably, highlighting challenges in processing and interpreting instructions presented purely through vision. Finally, in the semantically aligned setting, we observe a clear modality gap: despite receiving isomorphic representations, most models exhibit poorer calibration when reasoning with visual inputs compared to text. This gap persists even among strong models, with the exception of those explicitly trained for visual reasoning. These findings collectively suggest that while verbalized uncertainty is promising for interpretable confidence estimation, its reliability in VLMs is highly sensitive to both modality and model design, underscoring the need for modality-aware training and evaluation strategies. Comprehensive results for additional models and various prompts on the employed benchmark are provided in Appendix \S B and \S D.

\begin{table*}[htbp]
\centering
\resizebox{0.8\textwidth}{!}{
\tiny{
\begin{tabular}{ll*{5}{c}}
\toprule
Metric & Model & \multicolumn{5}{c}{IsoBench} \\
\cmidrule(lr){3-7} 
& & Mathematics & Games & Science & Algorithms & All \\
\midrule
\multirow{6}{*}{Acc $\uparrow$}
& Qwen2.5-VL 7B & 54.0 & 28.4 & 78.0 & 44.0 & 47.7 \\
& + Top-K & 57.5 & 28.1 & 73.3 & 47.1 & 49.6 \\
& + Self-Reflection & 52.7 & 27.7 & 78.0 & 47.6 & 47.7 \\
& + VCAP (Ours) & 55.2 & 30.0 & 80.0 & 46.1 & 49.2\\
& Qwen2.5-VL 72B & 60.3 & 24.1 & 86.0 & 61.2 & 53.7 \\
& + Top-K & 56.5 & 24.3 & 86.0 & 56.8 & 51.1 \\
& + Self-Reflection & 60.4 & 23.4 & 84.0 & 55.5 & 52.3 \\
& + VCAP (Ours) & 66.1 & 26.0 & 86.7 & 60.9 & 57.0 \\
\midrule
\multirow{6}{*}{ECE $\downarrow$}
& Qwen2.5-VL 7B & 0.377 & 0.695 & 0.155 & 0.526 & 0.467 \\
& + Top-K & 0.365 & 0.702 & 0.231 & 0.503 & 0.462 \\
& + Self-Reflection & 0.391 & 0.651 & 0.178 & 0.407 & 0.436 \\
& + VCAP (Ours) & 0.320 & 0.668 & 0.131 & 0.490 & 0.424 \\
& Qwen2.5-VL 72B & 0.353 & 0.724 & 0.146 & 0.376 & 0.431 \\
& + Top-K & 0.380 & 0.711 & 0.130 & 0.415 & 0.447 \\
& + Self-Reflection & 0.348 & 0.612 & 0.171 & 0.366 & 0.402 \\
& + VCAP (Ours) & 0.261 & 0.664 & 0.128 & 0.343 & 0.365 \\
\bottomrule
\end{tabular}
}}
\caption{IsoBench results across different categories and models with different prompting strategies. Image/text modality results are shown with slash (/). For Mathematics, the text modality shows LaTeX format results; for Games, the text modality shows PGN format results. All accuracy (Acc) values are in percentage.
}
\label{tab:diff_mod}
\vspace{-3mm}
\end{table*}

\section{\textsc{Visual Confidence-Aware Prompting (VCAP)}}
\begin{figure}[h]
    \centering
    \includegraphics[width=\columnwidth]{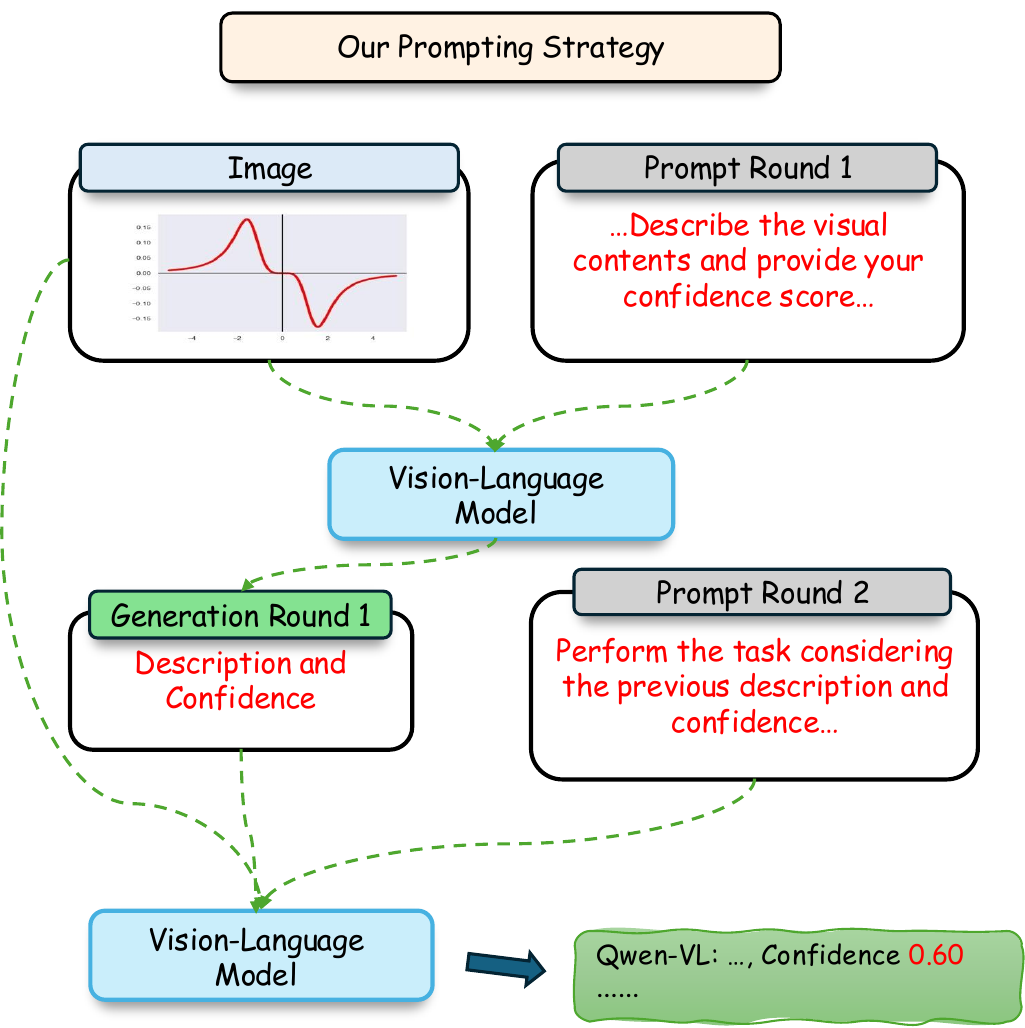}
    \caption{The illustration of our Visual Confidence-Aware Prompting (VCAP).}
    \label{fig:our_prompting}
    \vspace{-6mm}
\end{figure}

As shown in our evaluation results, miscalibration is a persistent issue across modalities in many VLMs, particularly among instruction-tuned models. To address this, we introduce a prompting strategy, Visual Confidence-Aware Prompting (VCAP), that leverages the multi-turn dialogue capabilities of instruction models to enhance calibration. Motivated by the observed modality gaps, we ask whether confidence signals derived from the vision modality can be explicitly incorporated into the final response generation. This leads us to design a two-stage prompting approach that encourages the model to separately reflect on visual confidence before producing a calibrated answer, as shown in \autoref{fig:our_prompting}.

VCAP separates visual understanding from task execution to improve calibration accuracy. In the first round, the VLM is asked to describe the visual input in detail and provide a confidence score, focusing exclusively on the visual modality to minimize cognitive load and isolate perception. In the second round, the model is prompted to complete the task and generate a verbalized confidence score, this time taking into account its prior self-assessment from the visual description. By decoupling perception and reasoning in a structured dialogue, the strategy encourages more reflective processing and aims to improve calibration, particularly on the visual modality side.

We evaluate our proposed approach on the IsoBench benchmark using the Qwen2.5-VL series. IsoBench consists of semantically aligned tasks presented in both visual and textual modalities, making it a suitable testbed for analyzing modality-specific calibration behavior. In the general evaluation setting, we adopt CoT prompting to elicit both reasoning steps and confidence estimates within a single turn, which we adopt as a baseline here. As for a stronger baseline, we compare against Top-K prompting, where the model generates multiple candidate answers with associated confidence scores, and the one with the highest confidence is selected as the final output \cite{tian-etal-2023-just, xiong2024llmsexpressuncertaintyempirical}. Additionally, we report results using self-reflection prompting \cite{xiong2024llmsexpressuncertaintyempirical, 10.1162/tacl_a_00737}, another two-round strategy in which the model first generates an answer, followed by a second prompt that elicits a confidence estimate for that response.

Results are shown in \autoref{tab:diff_mod}. Across both model sizes of Qwen2.5-VL, our two-stage prompting method leads to a moderate improvement in accuracy and a more notable gain in calibration performance, even when compared to the Top-K and self-reflection baselines. These findings suggest that explicitly structuring the confidence elicitation process through multi-round prompting in isolated modalities can help mitigate miscalibration and enhance the reliability of verbalized uncertainty in VLMs.

\section{Discussion}
In this paper, we first evaluate whether VLMs can express their uncertainties in a calibrated manner. Across various types of models and datasets, our results highlight the widespread miscalibration issue in current VLMs and suggest that vision-based reasoning can significantly improve both multimodal reasoning and reduce the modality gap. Building on this insight, we propose \textsc{Visual Confidence-aware Prompting}, a two-stage prompting strategy that explicitly guides VLMs to express more calibrated confidence by decoupling visual interpretation from task execution.

Previous research on verbalized uncertainty has typically focused on isolated aspects of instruct VLMs' behaviors. \citet{groot-valdenegro-toro-2024-overconfidence} evaluated verbalized uncertainty using a small 39-image Japanese-language dataset. \citet{borszukovszki-etal-2025-know} examined how VLMs respond to input noise, while \citet{zhao2025objectlevelverbalizedconfidencecalibration} investigated object-level miscalibration and proposed a two-stage fine-tuning approach for calibration. Our work expands upon these efforts by conducting a broader evaluation across multiple domains, prompting strategies, and modality configurations, showing that miscalibration is a consistent challenge in VLMs.

In parallel, the emergence of text reasoning models has reshaped the landscape of LLM development. \citet{zeng2025reasoningmodelsbetterverbalized} systematically evaluated verbalized uncertainty in reasoning models and found that they tend to produce more calibrated outputs than instruction-tuned models. Our study provides complementary evidence from a multimodal perspective, showing that modality-specific reasoning in VLMs, particularly reasoning grounded in visual input, contributes to improved calibration and confidence reliability.

Previous works have introduced a variety of fine-tuning and prompting strategies to improve verbalized uncertainty in LLMs. These include methods such as distilling self-consistency signals into the model \cite{hager2025uncertaintydistillationteachinglanguage} and using Top-K prompting to elicit more calibrated confidence scores \cite{tian-etal-2023-just, xiong2024llmsexpressuncertaintyempirical}. In this work, we extend these efforts to the multimodal domain by evaluating and improving verbalized calibration in VLMs. We propose a two-stage prompting strategy that first isolates visual understanding and then guides task execution based on self-assessed visual confidence. Our results show that this approach can effectively enhance calibration in VLMs across modalities. These findings underscore the importance of explicitly leveraging multimodal inputs and modality-specific reasoning when designing strategies for improving confidence estimation in VLMs. This line of work highlights the need for calibration-aware prompting designs that are sensitive to the structure and strengths of different input modalities.

\section*{Limitations}
In this work, we comprehensively evaluated how VLMs express uncertainty through natural language and proposed visual confidence-aware prompting to address the identified challenges. One key limitation is that although "think with images" models like o3 and o4-mini demonstrate exceptional verbalized calibration, their closed-source nature and lack of publicly available implementation details make it difficult to understand and analyze why they significantly outperform text reasoning and instruction-tuned models. Furthermore, since there are currently only two closed-source OpenAI models with production-level performance in the "think with images" category, we cannot investigate how different model families perform in terms of verbalized calibration within this approach. Additionally, while our goal was to evaluate verbalized calibration across a broad spectrum of tasks, we did not conduct dedicated experiments to investigate some of the challenging scenarios frequently encountered in downstream applications, such as temporal visual grounding in video understanding. We also consider these directions to be meaningful avenues for future work.

\bibliography{custom}

\newpage
\appendix

\section{Prompts}
\label{sec:appendix}
In this section, we provide all prompts used in this work.

\subsection{CoT prompting (vanilla)}
Here, we present the prompts used in the CoT prompting.

\subsubsection{IsoBench}
\begin{tcolorbox}[remarkbox, title=graph\_maxflow\_image\_vanilla]
\small
You are given an image of a graph and two query nodes. (one source node and one sink node). \space \\
The source node is the node where the flow starts and the sink node is the node where the flow ends.\\
\\
YOUR TASK is to solve the maxflow problem given the weighted directed graph and provide a confidence score (0\% to 100\%) for your answer.\\
\\
\textbf{Definition of Maxflow problem:}\\
In the max flow problem, we have a directed graph with a source node s and a sink node t, and each edge has a capacity (integer valued, colored in green) that represents the maximum amount of flow that can be sent through it.\\
The goal is to find the maximum amount of flow that can be sent from s to t, while respecting the capacity constraints on the edges.\\
\\
\textbf{Query Example:}\\
Source node (zero-indexed): 0\\
Sink node (zero-indexed): 2\\
In the query example, the nodes are zero-indexed.\\
\\
\textbf{Instructions:}\\
- Please reason step by step\\
- At the end, present your final answer and a confidence score in the following XML format:\\
<answer>[the maximum flow from the source node to the sink node (in Arabic digits)]</answer>\\
<confidence>[your confidence score for the answer]</confidence>\\
\\
\textbf{Example output:}\\
\char91 YOUR\_REASONING\char93\\
<answer>12</answer>\\
<confidence>80\%</confidence>\\
\end{tcolorbox}

\begin{tcolorbox}[remarkbox, title=graph\_maxflow\_text\_vanilla]
\small
You are given an adjacency matrix of a graph and two query nodes. (one source node and one sink node). \space \\
The source node is the node where the flow starts and the sink node is the node where the flow ends.\\
\\
YOUR TASK is to solve the maxflow problem given the weighted directed graph and provide a confidence score (0\% to 100\%) for your answer.\\
\\
\textbf{Definition of Maxflow problem:}\\
In the max flow problem, we have a directed graph with a source node s and a sink node t, and each edge has a capacity that represents the maximum amount of flow that can be sent through it.\\
The goal is to find the maximum amount of flow that can be sent from s to t, while respecting the capacity constraints on the edges.\\
\\
\textbf{Query Example:}\\
adjacency matrix:\\
\char91 0, 1, 4\char93\\
\char91 0, 0, 6\char93\\
\char91 0, 0, 0\char93\\
Source node (zero-indexed): 0\\
Sink node (zero-indexed): 2\\
In the query example, the nodes are zero-indexed.\\
\\
\textbf{Instructions:}\\
- Please reason step by step\\
- At the end, present your final answer and a confidence score in the following XML format:\\
<answer>[the maximum flow from the source node to the sink node (in Arabic digits)]</answer>\\
<confidence>[your confidence score for the answer]</confidence>\\
\\
\textbf{Example output:}\\
\char91 YOUR\_REASONING\char93\\
<answer>12</answer>\\
<confidence>80\%</confidence>\\
\end{tcolorbox}

\begin{tcolorbox}[remarkbox, title=graph\_connectivity\_image\_vanilla]
\small
You are given an image of a graph and two query nodes. \space \\
\\
YOUR TASK is to determine whether the query nodes are connected as True or False, and provide a confidence score (0\% to 100\%) for your prediction.\\
\\
\textbf{Query Example:}\\
Query node 1 (zero-indexed): 9\\
Query node 2 (zero-indexed): 4\\
In the query example, the nodes are zero-indexed.\\
\\
\textbf{Instructions:}\\
- Please reason step by step\\
- At the end, present your final answer and a confidence score in the following XML format:\\
<answer>[whether the query nodes are connected: "True" or "False"]</answer>\\
<confidence>[your confidence score for the answer]</confidence>\\
\\
\textbf{Example output:}\\
\char91 YOUR\_REASONING\char93\\
<answer>True</answer>\\
<confidence>80\%</confidence>\\
\end{tcolorbox}

\begin{tcolorbox}[remarkbox, title=graph\_connectivity\_text\_vanilla]
\small
You are given the adjacency matrix of a graph and two query nodes. \space \\
\\
YOUR TASK is to determine whether the query nodes are connected as True or False, and provide a confidence score (0\% to 100\%) for your prediction.\\
\\
\textbf{Query Example:}\\
adjacency matrix:\\
\char91 0, 0, 0, 0, 0, 0, 0, 0, 0, 0, 0, 0\char93\\
\char91 0, 0, 0, 0, 0, 0, 0, 0, 0, 0, 0, 0\char93\\
\char91 0, 0, 0, 0, 0, 0, 0, 0, 0, 0, 0, 0\char93\\
\char91 0, 0, 0, 0, 0, 0, 0, 0, 0, 0, 0, 0\char93\\
\char91 0, 0, 0, 0, 0, 0, 0, 0, 0, 0, 0, 0\char93\\
\char91 0, 0, 0, 0, 0, 0, 0, 0, 0, 0, 0, 0\char93\\
\char91 0, 0, 0, 0, 0, 0, 0, 0, 0, 0, 0, 0\char93\\
\char91 0, 0, 0, 0, 0, 0, 0, 0, 0, 0, 0, 0\char93\\
\char91 0, 0, 0, 0, 0, 0, 0, 0, 0, 0, 1, 0\char93\\
\char91 0, 0, 0, 0, 0, 0, 0, 0, 0, 0, 0, 0\char93\\
\char91 0, 0, 0, 0, 0, 0, 0, 0, 1, 0, 0, 0\char93\\
\char91 0, 0, 0, 0, 0, 0, 0, 0, 0, 0, 0, 0\char93\\
Query node 1 (zero-indexed): 9\\
Query node 2 (zero-indexed): 4\\
In the query example, the nodes are zero-indexed.\\
\\
\textbf{Instructions:}\\
- Please reason step by step\\
- At the end, present your final answer and a confidence score in the following XML format:\\
<answer>[whether the query nodes are connected: "True" or "False"]</answer>\\
<confidence>[your confidence score for the answer]</confidence>\\
\\
\textbf{Example output:}\\
\char91 YOUR\_REASONING\char93\\
<answer>True</answer>\\
<confidence>80\%</confidence>\\
\end{tcolorbox}

\begin{tcolorbox}[remarkbox, title=graph\_isomorphism\_image\_vanilla]
\small
You are given an image of two specific graphs, G (Left Graph) and H (Right Graph).\\
\\
YOUR TASK is to determine if graph G and graph H are \textbf{isomorphic} based on the image, and provide a confidence score (0\% to 100\%) for your determination.\\
\\
\textbf{Instructions:}\\
- Please reason step by step\\
- At the end, present your final answer and a confidence score in the following XML format:\\
<answer>[whether the two graphs are isomorphic: "True" or "False"]</answer>\\
<confidence>[your confidence score for the answer]</confidence>\\
\\
\textbf{Example output:}\\
\char91 YOUR\_REASONING\char93\\
<answer>True</answer>\\
<confidence>80\%</confidence>\\
\end{tcolorbox}

\begin{tcolorbox}[remarkbox, title=graph\_isomorphism\_text\_vanilla]
\small
You are given the adjacency matrix representations of two specific graphs, G and H.\\
\\
YOUR TASK is to determine if graph G and graph H, defined below, are \textbf{isomorphic} based on their provided adjacency matrices, and provide a confidence score (0\% to 100\%) for your determination.\\
\\
\textbf{Query Example:}\\
adjacency matrix G:\\
\char91 0, 0, 0, 0, 0, 0, 0, 0, 0, 0, 0, 0\char93\\
\char91 0, 0, 0, 0, 0, 0, 0, 0, 0, 0, 0, 0\char93\\
\char91 0, 0, 0, 0, 0, 0, 0, 0, 0, 0, 0, 0\char93\\
\char91 0, 0, 0, 0, 0, 0, 0, 0, 0, 0, 0, 0\char93\\
\char91 0, 0, 0, 0, 0, 0, 0, 0, 0, 0, 0, 0\char93\\
\char91 0, 0, 0, 0, 0, 0, 0, 0, 0, 0, 0, 0\char93\\
\char91 0, 0, 0, 0, 0, 0, 0, 0, 0, 0, 0, 0\char93\\
\char91 0, 0, 0, 0, 0, 0, 0, 0, 0, 0, 0, 0\char93\\
\char91 0, 0, 0, 0, 0, 0, 0, 0, 0, 0, 1, 0\char93\\
\char91 0, 0, 0, 0, 0, 0, 0, 0, 0, 0, 0, 0\char93\\
\char91 0, 0, 0, 0, 0, 0, 0, 0, 1, 0, 0, 0\char93\\
\char91 0, 0, 0, 0, 0, 0, 0, 0, 0, 0, 0, 0\char93\\
adjacency matrix H:\\
\char91 0, 0, 0, 0, 0, 0, 0, 0, 0, 0, 0, 0\char93\\
\char91 0, 0, 0, 0, 0, 0, 0, 0, 0, 0, 0, 0\char93\\
\char91 0, 0, 0, 0, 0, 0, 0, 0, 0, 0, 0, 0\char93\\
\char91 0, 0, 0, 0, 0, 0, 0, 0, 0, 0, 0, 0\char93\\
\char91 0, 0, 0, 0, 0, 0, 0, 0, 0, 0, 0, 0\char93\\
\char91 0, 0, 0, 0, 0, 0, 0, 0, 0, 0, 0, 0\char93\\
\char91 0, 0, 0, 0, 0, 0, 0, 0, 0, 0, 0, 0\char93\\
\char91 0, 0, 0, 0, 0, 0, 0, 0, 0, 1, 0, 0\char93\\
\char91 0, 0, 0, 0, 0, 0, 0, 0, 0, 0, 0, 0\char93\\
\char91 0, 0, 0, 0, 0, 0, 0, 1, 0, 0, 0, 0\char93\\
\char91 0, 0, 0, 0, 0, 0, 0, 0, 0, 0, 0, 0\char93\\
\char91 0, 0, 0, 0, 0, 0, 0, 0, 0, 0, 0, 0\char93\\
\\
\textbf{Instructions:}\\
- Please reason step by step\\
- At the end, present your final answer and a confidence score in the following XML format:\\
<answer>[whether the two graphs are isomorphic: "True" or "False"]</answer>\\
<confidence>[your confidence score for the answer]</confidence>\\
\\
\textbf{Example output:}\\
\char91 YOUR\_REASONING\char93\\
<answer>True</answer>\\
<confidence>80\%</confidence>\\
\end{tcolorbox}

\begin{tcolorbox}[remarkbox, title=puzzle\_image\_vanilla]
\small
You are given a visual representation of a chess puzzle for which a sequence of unique best moves is determinable (e.g. sequences of moves leading to a forced checkmate).\\
\\
\textbf{Definition of the Chess Puzzle:}\\
In a chess puzzle, you are required to make a series of optimal moves leading to checkmate, starting from the given position.\\
\\
YOUR TASK is to predict THE FIRST MOVE that should be played given this board setup, and provide a confidence score (0\% to 100\%) for your answer.\\
Your answer should specify the move in Algebraic Coordinate Notation (e.g., "d2d1", "e5a1", "c4f4").\\
\\
\textbf{Instructions:}\\
- Please reason step by step\\
- At the end, present your final answer and a confidence score in the following XML format:\\
<answer>[only the first move in Algebraic Coordinate Notation]</answer>\\
<confidence>[your confidence score for the answer]</confidence>\\
\\
\textbf{Example output:}\\
\char91 YOUR\_REASONING\char93\\
<answer>e2e4</answer>\\
<confidence>80\%</confidence>\\
\end{tcolorbox}

\begin{tcolorbox}[remarkbox, title=puzzle\_pgn\_vanilla]
\small
You are given a PGN representation of a chess puzzle for which a sequence of unique best moves is determinable (e.g. sequences of moves leading to a forced checkmate).\\
\\
\textbf{Definition of the Chess Puzzle:}\\
In a chess puzzle, you are required to make a series of optimal moves leading to checkmate, starting from the given position.\\
\\
YOUR TASK is to predict THE FIRST MOVE that should be played given this board setup, and provide a confidence score (0\% to 100\%) for your answer.\\
Your answer should specify the move in Algebraic Coordinate Notation (e.g., "d2d1", "e5a1", "c4f4").\\
\\
PGN: 1. e4 e6 2. d4 Ne7 3. c4 Ng6 4. Nf3 Nh4 5. Nxh4 Qxh4 6. Bd3 b6 7. O-O Bb7 8. Nc3 Nc6 9. d5 Ne7 10. Qf3 Ng6 11. Qg3 Qxg3 12. fxg3 Ne5 13. Be2 Bc5+ 14. Kh1 O-O 15. Bf4 Bd4 16. Rad1 Bxc3 17. bxc3 Ng6 18. Bxc7 exd5 19. cxd5 Rfe8 20. Bf3 Ne5 21. Bxe5 Rxe5 22. c4 Ba6 23. Rc1 d6 24. Rfe1 Rae8 25. Kg1 R8e7 26. Kf2 f5 27. exf5 Rxe1 28. Rxe1 Rxe1 29. Kxe1 Bxc4 30. a3 a5 31. Kd2 Kf7 32. Kc3 Bf1 33. h4 Kf6 34. g4 Ke5 35. h5 h6 36. Kb3 Kd4 37. Ka4 Bc4 38. g3 Ba6 39. g5 hxg5 40. f6 gxf6 41. h6 Bd3 42. g4 Kc5 43. Be2 Bh7 44. Bb5 Kxd5 45. Bd7 Bg8 46. Bf5 Ke5 47. h7 Bxh7 48. Bxh7 d5 49. Kb5 d4 50. Kc4 a4 51. Bc2 b5+ 52. Kxb5 Kf4 53. Bd1 d3 54. Kxa4 f5\\
\\
\textbf{Instructions:}\\
- Please reason step by step\\
- At the end, present your final answer and a confidence score in the following XML format:\\
<answer>[only the first move in Algebraic Coordinate Notation]</answer>\\
<confidence>[your confidence score for the answer]</confidence>\\
\\
\textbf{Example output:}\\
\char91 YOUR\_REASONING\char93\\
<answer>e2e4</answer>\\
<confidence>80\%</confidence>\\
\end{tcolorbox}

\begin{tcolorbox}[remarkbox, title=winner\_id\_image\_vanilla]
\small
You are given a visual representation of a chess puzzle for which a sequence of unique best moves is determinable (e.g. sequences of moves leading to a forced checkmate).\\
\\
\textbf{Definition of the Chess Puzzle:}\\
In a chess puzzle, you are required to make a series of optimal moves leading to checkmate, starting from the given position.\\
\\
YOUR TASK is to identify the winner of this game given this board setup, and provide a confidence score (0\% to 100\%) for your prediction.\\
Your answer should specify the winner as one of the following strings: "White", "Black", or "Draw".\\
\\
\textbf{Instructions:}\\
- Please reason step by step\\
- At the end, present your final answer and a confidence score in the following XML format:\\
<answer>[only the winner of this game: "White", "Black", or "Draw"]</answer>\\
<confidence>[your confidence score for the answer]</confidence>\\
\\
\textbf{Example output:}\\
\char91 YOUR\_REASONING\char93\\
<answer>Draw</answer>\\
<confidence>80\%</confidence>\\
\end{tcolorbox}

\begin{tcolorbox}[remarkbox, title=winner\_id\_pgn\_vanilla]
\small
You are given a PGN representation of a chess puzzle for which a sequence of unique best moves is determinable (e.g. sequences of moves leading to a forced checkmate).\\
\\
\textbf{Definition of the Chess Puzzle:}\\
In a chess puzzle, you are required to make a series of optimal moves leading to checkmate, starting from the given position.\\
\\
YOUR TASK is to identify the winner of this game given this board setup, and provide a confidence score (0\% to 100\%) for your prediction.\\
Your answer should specify the winner as one of the following strings: "White", "Black", or "Draw".\\
\\
PGN: 1. d4 d5 2. e3 e6 3. Bd3 Nf6 4. Nd2 Be7 5. c3 O-O 6. f4 Nbd7 7. Qe2 c5 8. Ngf3 c4 9. Bc2 a6 10. O-O b5 11. Ne5 Bb7 12. a3 Rb8 13. e4 dxe4 14. Nxe4 Nxe5 15. fxe5 Nd5 16. Qg4 a5 17. Bh6 f6 18. Qxg7\#\\
\\
\textbf{Instructions:}\\
- Please reason step by step\\
- At the end, present your final answer and a confidence score in the following XML format:\\
<answer>[only the winner of this game: "White", "Black", or "Draw"]</answer>\\
<confidence>[your confidence score for the answer]</confidence>\\
\\
\textbf{Example output:}\\
\char91 YOUR\_REASONING\char93\\
<answer>Draw</answer>\\
<confidence>80\%</confidence>\\
\end{tcolorbox}

\begin{tcolorbox}[remarkbox, title=image\_math\_parity\_vanilla]
\small
You are given a plot of a real-valued, scalar function f(x).\\
YOUR TASK is to determine whether f(x) is an even function, an odd function, or neither, and provide a confidence score (0\% to 100\%) for your answer.\\
- Definition of an even function: A function such that f(x) = f(-x) where the value remains unchanged if the sign of the independent variable is reversed.\\
- Definition of an odd function: A function such that f(-x) = -f(x) where the sign is reversed but the absolute value remains the same if the sign of the independent variable is reversed\\
- A function is neither even nor odd if it does not satisfy either definitions.\\
\\
\textbf{Instructions:}\\
- Please reason step by step\\
- At the end, present your final answer and a confidence score in the following XML format:\\
<answer>[only the final result: 'even', 'odd', or 'neither']</answer>\\
<confidence>[your confidence score for the answer]</confidence>\\
\\
\textbf{Example output:}\\
\char91 YOUR\_REASONING\char93\\
<answer>even</answer>\\
<confidence>80\%</confidence>\\
\end{tcolorbox}

\begin{tcolorbox}[remarkbox, title=image\_math\_convexity\_vanilla]
\small
You are given a plot of a real-valued, scalar function f(x).\\
YOUR TASK is to determine whether f(x) is an convex function or a concave function and provide a confidence score (0\% to 100\%) for your answer\\
- Definition of a convex function: A function such that for all x, y, and $0 \leq t \leq 1$\\
$f(tx + (1 - t)y) \leq tf(x) + (1 - t)f(y)$\\
- Definition of a concave function: A function such that for all x, y, and $0 \leq t \leq 1$\\
$f(tx + (1 - t)y) \geq tf(x) + (1 - t)f(y)$\\
\\
\textbf{Instructions:}\\
- Please reason step by step\\
- At the end, present your final answer and a confidence score in the following XML format:\\
<answer>[only the final result: 'convex' or 'concave']</answer>\\
<confidence>[your confidence score for the answer]</confidence>\\
\\
\textbf{Example output:}\\
\char91 YOUR\_REASONING\char93\\
<answer>convex</answer>\\
<confidence>80\%</confidence>\\
\end{tcolorbox}

\begin{tcolorbox}[remarkbox, title=image\_math\_breakpoint\_vanilla]
\small
You are given a plot of a real-valued, scalar function f(x).\\
YOUR TASK is to count the number of breakpoints in the plot of f(x) and provide a confidence score (0\% to 100\%) for your answer. \space \\
A breakpoint refers to a point on the function’s domain at which the function changes its slope.\\
\\
You should IGNORE the left and right end point of the domain, i.e. if the function is defined on [a, b], you should only consider the domain (a, b).\\
\\
\textbf{Instructions:}\\
- Please reason step by step\\
- At the end, present your final answer and a confidence score in the following XML format:\\
<answer>[the number of breakpoints (in Arabic digits)]</answer>\\
<confidence>[your confidence score for the answer]</confidence>\\
\\
\textbf{Example output:}\\
\char91 YOUR\_REASONING\char93\\
<answer>2</answer>\\
<confidence>80\%</confidence>\\
\end{tcolorbox}

\begin{tcolorbox}[remarkbox, title=text\_math\_parity\_vanilla]
\small
You are given a real-valued, scalar function f(x).\\
YOUR TASK is to determine whether f(x) is an even function, an odd function, or neither, and provide a confidence score (0\% to 100\%) for your answer.\\
- Definition of an even function: A function such that f(x) = f(-x) where the value remains unchanged if the sign of the independent variable is reversed.\\
- Definition of an odd function: A function such that f(-x) = -f(x) where the sign is reversed but the absolute value remains the same if the sign of the independent variable is reversed\\
- A function is neither even nor odd if it does not satisfy either definitions.\\
\\
Here is the expression of f(x){domain}:\\
\{text\}\\
\\
\textbf{Instructions:}\\
- Please reason step by step\\
- At the end, present your final answer and a confidence score in the following XML format:\\
<answer>[only the final result: 'even', 'odd', or 'neither']</answer>\\
<confidence>[your confidence score for the answer]</confidence>\\
\\
\textbf{Example output:}\\
\char91 YOUR\_REASONING\char93\\
<answer>even</answer>\\
<confidence>80\%</confidence>\\
\end{tcolorbox}

\begin{tcolorbox}[remarkbox, title=text\_math\_convexity\_vanilla]
\small
You are given a real-valued, scalar function f(x).\\
YOUR TASK is to determine whether f(x) is an convex function or a concave function and provide a confidence score (0\% to 100\%) for your answer\\
- Definition of a convex function: A function such that for all x, y, and $0 \leq t \leq 1$\\
$f(tx + (1 - t)y) \leq tf(x) + (1 - t)f(y)$\\
- Definition of a concave function: A function such that for all x, y, and $0 \leq t \leq 1$\\
$f(tx + (1 - t)y) \geq tf(x) + (1 - t)f(y)$\\
\\
Here is the expression of f(x){domain}:\\
\{text\}\\
\\
\textbf{Instructions:}\\
- Please reason step by step\\
- At the end, present your final answer and a confidence score in the following XML format:\\
<answer>[only the final result: 'convex' or 'concave']</answer>\\
<confidence>[your confidence score for the answer]</confidence>\\
\\
\textbf{Example output:}\\
\char91 YOUR\_REASONING\char93\\
<answer>convex</answer>\\
<confidence>80\%</confidence>\\
\end{tcolorbox}

\begin{tcolorbox}[remarkbox, title=text\_math\_breakpoint\_vanilla]
\small
You are given a real-valued, scalar function f(x).\\
YOUR TASK is to count the number of breakpoints in the plot of f(x) and provide a confidence score (0\% to 100\%) for your answer. \space \\
A breakpoint refers to a point on the function’s domain at which the function changes its slope.\\
\\
Here is the expression of f(x){domain}:\\
\{text\}\\
\\
You should IGNORE the left and right end point of the domain, i.e. if the function is defined on [a, b], you should only consider the domain (a, b).\\
\\
\textbf{Instructions:}\\
- Please reason step by step\\
- At the end, present your final answer and a confidence score in the following XML format:\\
<answer>[the number of breakpoints (in Arabic digits)]</answer>\\
<confidence>[your confidence score for the answer]</confidence>\\
\\
\textbf{Example output:}\\
\char91 YOUR\_REASONING\char93\\
<answer>2</answer>\\
<confidence>80\%</confidence>\\
\end{tcolorbox}

\begin{tcolorbox}[remarkbox, title=chemistry\_image\_vanilla]
\small
You are given an image of a chemistry diagram.\\
YOUR TASK is to read the question and select the correct answer from the provided options and provide a confidence score (0\% to 100\%) for your answer.\\
\\
Which solution has a higher concentration of green particles?\\
A. Solution B\\
B. neither; their concentrations are the same\\
C. Solution A\\
\\
\textbf{Instructions:}\\
- Carefully read and analyze the problem.\\
- Reason through the solution step by step, if helpful.\\
- At the end, present your final answer and a confidence score in the following XML format:\\
<answer>[only the letter of the correct answer]</answer>\\
<confidence>[your confidence score here]</confidence>\\
\\
\textbf{Example output:}\\
\char91 YOUR\_REASONING\char93\\
<answer>A</answer>\\
<confidence>80\%</confidence>\\
\end{tcolorbox}

\begin{tcolorbox}[remarkbox, title=chemistry\_text\_vanilla]
\small
You are given a multiple-choice chemistry question.\\
YOUR TASK is to read the question and select the correct answer from the provided options and provide a confidence score (0\% to 100\%) for your answer.\\
\\
In Solution A and Solution B, the green particles represent the solute. The volume of the solvent in two containers are equal. Solution A and Solution B have the same number of green particles.\\
\\
Which solution has a higher concentration of green particles?\\
A. Solution B\\
B. neither; their concentrations are the same\\
C. Solution A\\
\\
\textbf{Instructions:}\\
- Carefully read and analyze the problem.\\
- Reason through the solution step by step, if helpful.\\
- At the end, present your final answer and a confidence score in the following XML format:\\
<answer>[only the letter of the correct answer]</answer>\\
<confidence>[your confidence score here]</confidence>\\
\\
\textbf{Example output:}\\
\char91 YOUR\_REASONING\char93\\
<answer>A</answer>\\
<confidence>80\%</confidence>\\
\end{tcolorbox}

\begin{tcolorbox}[remarkbox, title=physics\_image\_vanilla]
\small
You are given an image of a physics diagram.\\
YOUR TASK is to read the question and select the correct answer from the provided options and provide a confidence score (0\% to 100\%) for your answer.\\
\\
During this time, thermal energy was transferred from () to ().\\
A. the surroundings . . . each salmon\\
B. each salmon . . . the surroundings\\
\\
\textbf{Instructions:}\\
- Carefully read and analyze the problem.\\
- Reason through the solution step by step, if helpful.\\
- At the end, present your final answer and a confidence score in the following XML format:\\
<answer>[only the letter of the correct answer]</answer>\\
<confidence>[your confidence score here]</confidence>\\
\\
\textbf{Example output:}\\
\char91 YOUR\_REASONING\char93\\
<answer>A</answer>\\
<confidence>80\%</confidence>\\
\end{tcolorbox}

\begin{tcolorbox}[remarkbox, title=physics\_text\_vanilla]
\small
You are given a multiple-choice physics question.\\
YOUR TASK is to read the question and select the correct answer from the provided options and provide a confidence score (0\% to 100\%) for your answer.\\
\\
The temperature of each salmon increased.\\
\\
During this time, thermal energy was transferred from () to ().\\
A. the surroundings . . . each salmon\\
B. each salmon . . . the surroundings\\
\\
\textbf{Instructions:}\\
- Carefully read and analyze the problem.\\
- Reason through the solution step by step, if helpful.\\
- At the end, present your final answer and a confidence score in the following XML format:\\
<answer>[only the letter of the correct answer]</answer>\\
<confidence>[your confidence score here]</confidence>\\
\\
\textbf{Example output:}\\
\char91 YOUR\_REASONING\char93\\
<answer>A</answer>\\
<confidence>80\%</confidence>\\
\end{tcolorbox}

\subsubsection{MMMU-Pro}
\begin{tcolorbox}[remarkbox, title=standard\_10\_vanilla]
\small
\{question\}\\
A. \{option1\}\\
B. \{option2\}\\
C. \{option3\}\\
D. \{option4\}\\
E. \{option5\}\\
F. \{option6\}\\
G. \{option7\}\\
H. \{option8\}\\
I. \{option9\}\\
J. \{option10\}\\
\\
Answer the preceding multiple choice question and provide a confidence score (0\% to 100\%) for your answer.\\
\\
\textbf{Instructions:}\\
- Carefully read and analyze the problem.\\
- Reason through the solution step by step, if helpful.\\
- At the end, present your final answer and a confidence score in the following XML format:\\
<answer>[only the letter of the correct answer]</answer>\\
<confidence>[your confidence score here]</confidence>\\
\\
\textbf{Example output:}\\
\char91 YOUR\_REASONING\char93\\
<answer>A</answer>\\
<confidence>80\%</confidence>\\
\end{tcolorbox}

\begin{tcolorbox}[remarkbox, title=standard\_4\_vanilla]
\small
\{question\}\\
A. \{option1\}\\
B. \{option2\}\\
C. \{option3\}\\
D. \{option4\}\\
\\
Answer the preceding multiple choice question and provide a confidence score (0\% to 100\%) for your answer.\\
\\
\textbf{Instructions:}\\
- Carefully read and analyze the problem.\\
- Reason through the solution step by step, if helpful.\\
- At the end, present your final answer and a confidence score in the following XML format:\\
<answer>[only the letter of the correct answer]</answer>\\
<confidence>[your confidence score here]</confidence>\\
\\
\textbf{Example output:}\\
\char91 YOUR\_REASONING\char93\\
<answer>A</answer>\\
<confidence>80\%</confidence>\\
\end{tcolorbox}

\begin{tcolorbox}[remarkbox, title=vision\_vanilla]
\small
Write out the multiple-choice question in the image and then solve it. Also, provide a confidence score (0\% to 100\%) for your answer.\\
\\
\textbf{Instructions:}\\
- Carefully read and analyze the problem.\\
- Reason through the solution step by step, if helpful.\\
- At the end, present your final answer and a confidence score in the following XML format:\\
<answer>[only the letter of the correct answer]</answer>\\
<confidence>[your confidence score here]</confidence>\\
\\
\textbf{Example output:}\\
\char91 YOUR\_REASONING\char93\\
<answer>A</answer>\\
<confidence>80\%</confidence>\\
\end{tcolorbox}

\subsubsection{VideoMMMU}

\begin{tcolorbox}[remarkbox, title=adaptation\_mcq\_vanilla]
\small
You should watch and learn the video content. Then apply what you learned to answer the following multi-choice question. The image for this question is at the end of the video.\\
\\
\{question\}\\
A. \{option1\}\\
B. \{option2\}\\
C. \{option3\}\\
D. \{option4\}\\
E. \{option5\}\\
F. \{option6\}\\
G. \{option7\}\\
H. \{option8\}\\
I. \{option9\}\\
J. \{option10\}\\
\\
\textbf{Instructions:}\\
- Carefully read and analyze the problem.\\
- Reason through the solution step by step, if helpful.\\
- At the end, present your final answer and a confidence score in the following XML format:\\
<answer>[only the letter of the correct answer]</answer>\\
<confidence>[your confidence score here]</confidence>\\
\\
\textbf{Example output:}\\
\char91 YOUR\_REASONING\char93\\
<answer>A</answer>\\
<confidence>80\%</confidence>\\
\end{tcolorbox}

\begin{tcolorbox}[remarkbox, title=adaptation\_oe\_vanilla]
\small
You should watch and learn the video content. Then apply what you learned to answer the following open-ended question. The image for this question is at the end of the video.\\
\\
\{question\}\\
\\
\textbf{Instructions:}\\
- Carefully read and analyze the problem.\\
- Reason through the solution step by step, if helpful.\\
- At the end, present your final answer and a confidence score in the following XML format:\\
<answer>[only the correct answer]</answer>\\
<confidence>[your confidence score here]</confidence>\\
\\
\textbf{Example output:}\\
\char91 YOUR\_REASONING\char93\\
<answer>12</answer>\\
<confidence>80\%</confidence>\\
\end{tcolorbox}

\begin{tcolorbox}[remarkbox, title=perception\_and\_comprehension\_vanilla]
\small
\{question\}\\
A. \{option1\}\\
B. \{option2\}\\
C. \{option3\}\\
D. \{option4\}\\
E. \{option5\}\\
F. \{option6\}\\
G. \{option7\}\\
H. \{option8\}\\
I. \{option9\}\\
J. \{option10\}\\
\\
Please ignore the Quiz question in last frame of the video.\\
\\
\textbf{Instructions:}\\
- Carefully read and analyze the problem.\\
- Reason through the solution step by step, if helpful.\\
- At the end, present your final answer and a confidence score in the following XML format:\\
<answer>[only the letter of the correct answer]</answer>\\
<confidence>[your confidence score here]</confidence>\\
\\
\textbf{Example output:}\\
\char91 YOUR\_REASONING\char93\\
<answer>A</answer>\\
<confidence>80\%</confidence>\\
\end{tcolorbox}

\subsubsection{Visual SimpleQA}

\begin{tcolorbox}[remarkbox, title=multimodal\_vanilla]
\small
\textbf{Task:} Solve the following QA problem based on the given image. Provide your best guess along with a confidence score (0\% to 100\%).\\
\\
\textbf{Instructions:}\\
- Carefully read and analyze the problem.\\
- Reason through the solution step by step.\\
- At the end, present your final answer and a confidence score in the following XML format:\\
<answer>your final answer here</answer>\\
<confidence>your confidence score here</confidence>\\
\\
\textbf{Example output:}\\
\char91 YOUR\_REASONING\char93\\
<answer>123</answer>\\
<confidence>80\%</confidence>\\
\\
Now, here is the problem:\\
\{problem\}
\end{tcolorbox}

\begin{tcolorbox}[remarkbox, title=text\_only\_vanilla]
\small
\textbf{Task:} Solve the following QA problem. Provide your best guess along with a confidence score (0\% to 100\%).\\
\\
\textbf{Instructions:}\\
- Carefully read and analyze the problem.\\
- Reason through the solution step by step.\\
- At the end, present your final answer and a confidence score in the following XML format:\\
<answer>your final answer here</answer>\\
<confidence>your confidence score here</confidence>\\
\\
\textbf{Example output:}\\
\char91 YOUR\_REASONING\char93\\
<answer>123</answer>\\
<confidence>80\%</confidence>\\
\\
Now, here is the problem:\\
\{problem\}
\end{tcolorbox}

\subsubsection{MathVista}
For MathVista, we follow the official prompt and capture the answer and confidence scores by the same XML tags.

\subsubsection{MathVision}
\begin{tcolorbox}[remarkbox, title=mcq\_vanilla]
\small
\{question\}\\
\textbf{Choices:}\\
A. \{option1\}\\
B. \{option2\}\\
C. \{option3\}\\
D. \{option4\}\\
E. \{option5\}\\
F. \{option6\}\\
G. \{option7\}\\
H. \{option8\}\\
I. \{option9\}\\
J. \{option10\}\\
\\
\textbf{Instructions:}\\
- Carefully read and analyze the problem.\\
- Reason through the solution step by step.\\
- At the end, present your final answer and a confidence score in the following XML format\\
<answer>[only the correct answer using a single word or phrase]</answer>\\
<confidence>[your confidence score here]</confidence>\\
\\
\textbf{Example output:}\\
\char91 YOUR\_REASONING\char93\\
<answer>A</answer>\\
<confidence>80\%</confidence>
\end{tcolorbox}

\begin{tcolorbox}[remarkbox, title=oe\_vanilla]
\small
\{question\}\\
\\
\textbf{Instructions:}\\
- Carefully read and analyze the problem.\\
- Reason through the solution step by step.\\
- At the end, present your final answer and a confidence score in the following XML format:\\
<answer>[only the correct answer]</answer>\\
<confidence>[your confidence score here]</confidence>\\
\\
\textbf{Example output:}\\
\char91 YOUR\_REASONING\char93\\
<answer>12</answer>\\
<confidence>80\%</confidence>\\
\end{tcolorbox}

\subsection{VCAP prompting (1st round)}

\subsubsection{IsoBench}

\begin{tcolorbox}[remarkbox, title=graph\_maxflow\_image\_vcap\_1st]
\small
You are given an image of a graph and two query nodes. (one source node and one sink node). \space \\
The source node is the node where the flow starts and the sink node is the node where the flow ends.\\
\\
YOUR TASK is to describe the image with enough detail, and provide a confidence score (0\% to 100\%) for your description.\\
\\
Please reason step by step.\\
\end{tcolorbox}

\begin{tcolorbox}[remarkbox, title=graph\_connectivity\_image\_vcap\_1st]
\small
You are given an image of a graph and two query nodes. \space \\
\\
YOUR TASK is to describe the image with enough detail, and provide a confidence score (0\% to 100\%) for your description.\\
\\
Please reason step by step.\\
\end{tcolorbox}

\begin{tcolorbox}[remarkbox, title=graph\_isomorphism\_image\_vcap\_1st]
\small
You are given an image of two specific graphs, G (Left Graph) and H (Right Graph).\\
\\
YOUR TASK is to describe the image with enough detail, and provide a confidence score (0\% to 100\%) for your description.\\
\\
Please reason step by step.\\
\end{tcolorbox}

\begin{tcolorbox}[remarkbox, title=puzzle\_image\_vcap\_1st]
\small
You are given a visual representation of a chess puzzle for which a sequence of unique best moves is determinable (e.g. sequences of moves leading to a forced checkmate).\\
\\
\textbf{Definition of the Chess Puzzle:}\\
In a chess puzzle, you are required to make a series of optimal moves leading to checkmate, starting from the given position.\\
\\
YOUR TASK is to describe the visual presentation with enough detail, and provide a confidence score (0\% to 100\%) for your description.\\
\\
Please reason step by step.\\
\end{tcolorbox}

\begin{tcolorbox}[remarkbox, title=winner\_id\_image\_vcap\_1st]
\small
You are given a visual representation of a chess puzzle for which a sequence of unique best moves is determinable (e.g. sequences of moves leading to a forced checkmate).\\
\\
\textbf{Definition of the Chess Puzzle:}\\
In a chess puzzle, you are required to make a series of optimal moves leading to checkmate, starting from the given position.\\
\\
YOUR TASK is to describe the visual presentation with enough detail, and provide a confidence score (0\% to 100\%) for your description.\\
\\
Please reason step by step.\\
\end{tcolorbox}

\begin{tcolorbox}[remarkbox, title=image\_math\_parity\_vcap\_1st]
\small
You are given a plot of a real-valued, scalar function f(x).\\
YOUR TASK is to describe the plot with enough detail, and provide a confidence score (0\% to 100\%) for your description.\\
\\
Please reason step by step.\\
\end{tcolorbox}

\begin{tcolorbox}[remarkbox, title=image\_math\_convexity\_vcap\_1st]
\small
You are given a plot of a real-valued, scalar function f(x).\\
YOUR TASK is to describe the plot with enough detail, and provide a confidence score (0\% to 100\%) for your description.\\
\\
Please reason step by step.\\
\end{tcolorbox}

\begin{tcolorbox}[remarkbox, title=image\_math\_breakpoint\_vcap\_1st]
\small
You are given a plot of a real-valued, scalar function f(x).\\
YOUR TASK is to describe the plot with enough detail, and provide a confidence score (0\% to 100\%) for your description.\\
\\
Please reason step by step.\\
\end{tcolorbox}

\begin{tcolorbox}[remarkbox, title=chemistry\_image\_vcap\_1st]
\small
You are given an image of a chemistry diagram.\\
YOUR TASK is to describe the image with enough detail, and provide a confidence score (0\% to 100\%) for your description.\\
\\
Which solution has a higher concentration of green particles?\\
A. Solution B\\
B. neither; their concentrations are the same\\
C. Solution A\\
\\
Please reason step by step.\\
\end{tcolorbox}

\begin{tcolorbox}[remarkbox, title=physics\_image\_vcap\_1st]
\small
You are given an image of a physics diagram.\\
YOUR TASK is to describe the image with enough detail, and provide a confidence score (0\% to 100\%) for your description.\\
\\
During this time, thermal energy was transferred from () to ().\\
A. the surroundings . . . each salmon\\
B. each salmon . . . the surroundings\\
\\
Please reason step by step.\\
\end{tcolorbox}

\subsection{VCAP prompting (2nd round)}
Here, we present the prompts used in the second round of VCAP.
\subsubsection{IsoBench}

\begin{tcolorbox}[remarkbox, title=graph\_maxflow\_image\_vcap\_2nd]
\small
You are given an image of a graph and two query nodes. (one source node and one sink node). \space \\
The source node is the node where the flow starts and the sink node is the node where the flow ends.\\
\\
You have generated the following description of the image with a confidence score:\\
\{description\}\\
\\
YOUR TASK is to solve the maxflow problem given the weighted directed graph (shown in the image and described in your description) and provide a confidence score (0\% to 100\%) for your answer.\\
\\
\textbf{Definition of Maxflow problem:}\\
In the max flow problem, we have a directed graph with a source node s and a sink node t, and each edge has a capacity (integer valued, colored in green) that represents the maximum amount of flow that can be sent through it.\\
The goal is to find the maximum amount of flow that can be sent from s to t, while respecting the capacity constraints on the edges.\\
\\
\textbf{Query Example:}\\
Source node (zero-indexed): 0\\
Sink node (zero-indexed): 2\\
In the query example, the nodes are zero-indexed.\\
\\
\textbf{Instructions:}\\
- Please reason step by step, considering both the image and your own description\\
- Take into account your confidence score of the description\\
- At the end, present your final answer and a confidence score in the following XML format:\\
<answer>[the maximum flow from the source node to the sink node (in Arabic digits)]</answer>\\
<confidence>[your confidence score for the answer]</confidence>\\
\\
\textbf{Example output:}\\
\char91 YOUR\_REASONING\char93\\
<answer>12</answer>\\
<confidence>80\%</confidence>\\
\end{tcolorbox}

\begin{tcolorbox}[remarkbox, title=graph\_connectivity\_image\_vcap\_2nd]
\small
You are given an image of a graph and two query nodes. \space \\
\\
You have generated the following description of the image with a confidence score:\\
\{description\}\\
\\
YOUR TASK is to determine whether the query nodes are connected as True or False (shown in the image and described in your description), and provide a confidence score (0\% to 100\%) for your prediction.\\
\\
\textbf{Query Example:}\\
Query node 1 (zero-indexed): 9\\
Query node 2 (zero-indexed): 4\\
In the query example, the nodes are zero-indexed.\\
\\
\textbf{Instructions:}\\
- Please reason step by step, considering both the image and your own description\\
- Take into account your confidence score of the description\\
- At the end, present your final answer and a confidence score in the following XML format:\\
<answer>[whether the query nodes are connected: "True" or "False"]</answer>\\
<confidence>[your confidence score for the answer]</confidence>\\
\\
\textbf{Example output:}\\
\char91 YOUR\_REASONING\char93\\
<answer>True</answer>\\
<confidence>80\%</confidence>\\
\end{tcolorbox}

\begin{tcolorbox}[remarkbox, title=graph\_isomorphism\_image\_vcap\_2nd]
\small
You are given an image of two specific graphs, G (Left Graph) and H (Right Graph).\\
\\
You have generated the following description of the image with a confidence score:\\
\{description\}\\
\\
YOUR TASK is to determine if graph G and graph H are \textbf{isomorphic} based on the image (shown in the image and described in your description), and provide a confidence score (0\% to 100\%) for your determination.\\
\\
\textbf{Instructions:}\\
- Please reason step by step, considering both the image and your own description\\
- Take into account your confidence score of the description\\
- At the end, present your final answer and a confidence score in the following XML format:\\
<answer>[whether the two graphs are isomorphic: "True" or "False"]</answer>\\
<confidence>[your confidence score for the answer]</confidence>\\
\\
\textbf{Example output:}\\
\char91 YOUR\_REASONING\char93\\
<answer>True</answer>\\
<confidence>80\%</confidence>\\
\end{tcolorbox}

\begin{tcolorbox}[remarkbox, title=image\_math\_parity\_vcap\_2nd]
\small
You are given a plot of a real-valued, scalar function f(x).\\
You have generated the following description of the plot with a confidence score:\\
{description}\\
\\
YOUR TASK is to determine whether f(x) is an even function, an odd function, or neither (shown in the image and described in your description), and provide a confidence score (0\% to 100\%) for your answer.\\
- Definition of an even function: A function such that f(x) = f(-x) where the value remains unchanged if the sign of the independent variable is reversed.\\
- Definition of an odd function: A function such that f(-x) = -f(x) where the sign is reversed but the absolute value remains the same if the sign of the independent variable is reversed\\
- A function is neither even nor odd if it does not satisfy either definitions.\\
\\
\textbf{Instructions:}\\
- Please reason step by step, considering both the image and your own description\\
- Take into account your confidence score of the description\\
- At the end, present your final answer and a confidence score in the following XML format:\\
<answer>[only the final result: 'even', 'odd', or 'neither']</answer>\\
<confidence>[your confidence score for the answer]</confidence>\\
\\
\textbf{Example output:}\\
\char91 YOUR\_REASONING\char93\\
<answer>even</answer>\\
<confidence>80\%</confidence>\\
\end{tcolorbox}

\begin{tcolorbox}[remarkbox, title=image\_math\_convexity\_vcap\_2nd]
\small
You are given a plot of a real-valued, scalar function f(x).\\
You have generated the following description of the plot with a confidence score:\\
{description}\\
\\
YOUR TASK is to determine whether f(x) is an convex function or a concave function (shown in the image and described in your description) and provide a confidence score (0\% to 100\%) for your answer\\
- Definition of a convex function: A function such that for all x, y, and $0 \leq t \leq 1$\\
$f(tx + (1 - t)y) \leq tf(x) + (1 - t)f(y)$\\
- Definition of a concave function: A function such that for all x, y, and $0 \leq t \leq 1$\\
$f(tx + (1 - t)y) \geq tf(x) + (1 - t)f(y)$\\
\\
\textbf{Instructions:}\\
- Please reason step by step, considering both the image and your own description\\
- Take into account your confidence score of the description\\
- At the end, present your final answer and a confidence score in the following XML format:\\
<answer>[only the final result: 'convex' or 'concave']</answer>\\
<confidence>[your confidence score for the answer]</confidence>\\
\\
\textbf{Example output:}\\
\char91 YOUR\_REASONING\char93\\
<answer>convex</answer>\\
<confidence>80\%</confidence>\\
\end{tcolorbox}

\begin{tcolorbox}[remarkbox, title=image\_math\_breakpoint\_vcap\_2nd]
\small
You are given a plot of a real-valued, scalar function f(x).\\
You have generated the following description of the plot with a confidence score:\\
{description}\\
\\
YOUR TASK is to count the number of breakpoints in the plot of f(x) (shown in the image and described in your description) and provide a confidence score (0\% to 100\%) for your answer. \space \\
A breakpoint refers to a point on the function’s domain at which the function changes its slope.\\
\\
You should IGNORE the left and right end point of the domain, i.e. if the function is defined on [a, b], you should only consider the domain (a, b).\\
\\
\textbf{Instructions:}\\
- Please reason step by step, considering both the image and your own description\\
- Take into account your confidence score of the description\\
- At the end, present your final answer and a confidence score in the following XML format:\\
<answer>[the number of breakpoints (in Arabic digits)]</answer>\\
<confidence>[your confidence score for the answer]</confidence>\\
\\
\textbf{Example output:}\\
\char91 YOUR\_REASONING\char93\\
<answer>2</answer>\\
<confidence>80\%</confidence>\\
\end{tcolorbox}

\begin{tcolorbox}[remarkbox, title=puzzle\_image\_vcap\_2nd]
\small
You are given a visual representation of a chess puzzle for which a sequence of unique best moves is determinable (e.g. sequences of moves leading to a forced checkmate).\\
\\
\textbf{Definition of the Chess Puzzle:}\\
In a chess puzzle, you are required to make a series of optimal moves leading to checkmate, starting from the given position.\\
\\
You have generated the following description of the visual representation with a confidence score:\\
{description}\\
\\
YOUR TASK is to predict THE FIRST MOVE that should be played given this board setup (shown in the image and described in your description), and provide a confidence score (0\% to 100\%) for your answer.\\
Your answer should specify the move in Algebraic Coordinate Notation (e.g., "d2d1", "e5a1", "c4f4").\\
\\
\textbf{Instructions:}\\
- Please reason step by step, considering both the image and your own description\\
- Take into account your confidence score of the description\\
- At the end, present your final answer and a confidence score in the following XML format:\\
<answer>[only the first move in Algebraic Coordinate Notation]</answer>\\
<confidence>[your confidence score for the answer]</confidence>\\
\\
\textbf{Example output:}\\
\char91 YOUR\_REASONING\char93\\
<answer>e2e4</answer>\\
<confidence>80\%</confidence>\\
\end{tcolorbox}

\begin{tcolorbox}[remarkbox, title=winner\_id\_image\_vcap\_2nd]
\small
You are given a visual representation of a chess puzzle for which a sequence of unique best moves is determinable (e.g. sequences of moves leading to a forced checkmate).\\
\\
\textbf{Definition of the Chess Puzzle:}\\
In a chess puzzle, you are required to make a series of optimal moves leading to checkmate, starting from the given position.\\
\\
You have generated the following description of the visual representation with a confidence score:\\
{description}\\
\\
YOUR TASK is to identify the winner of this game given this board setup (shown in the image and described in your description), and provide a confidence score (0\% to 100\%) for your answer.\\
Your answer should specify the winner as one of the following strings: "White", "Black", or "Draw".\\
\\
\textbf{Instructions:}\\
- Please reason step by step, considering both the image and your own description\\
- Take into account your confidence score of the description\\
- At the end, present your final answer and a confidence score in the following XML format:\\
<answer>[only the winner of this game: "White", "Black", or "Draw"]</answer>\\
<confidence>[your confidence score for the answer]</confidence>\\
\\
\textbf{Example output:}\\
\char91 YOUR\_REASONING\char93\\
<answer>Draw</answer>\\
<confidence>80\%</confidence>\\
\end{tcolorbox}

\begin{tcolorbox}[remarkbox, title=chemistry\_image\_vcap\_2nd]
\small
You are given an image of a chemistry diagram.\\
You have generated the following description of the image with a confidence score:\\
{description}\\
\\
YOUR TASK is to read the question and select the correct answer from the provided options (shown in the image and described in your description) and provide a confidence score (0\% to 100\%) for your answer.\\
\\
Which solution has a higher concentration of green particles?\\
A. Solution B\\
B. neither; their concentrations are the same\\
C. Solution A\\
\\
\textbf{Instructions:}\\
- Carefully read and analyze the problem.\\
- Reason through the solution step by step, if helpful, considering both the image and your own description\\
- Take into account your confidence score of the description\\
- At the end, present your final answer and a confidence score in the following XML format:\\
<answer>[only the letter of the correct answer]</answer>\\
<confidence>[your confidence score here]</confidence>\\
\\
\textbf{Example output:}\\
\char91 YOUR\_REASONING\char93\\
<answer>A</answer>\\
<confidence>80\%</confidence>\\
\end{tcolorbox}

\begin{tcolorbox}[remarkbox, title=physics\_image\_vcap\_2nd]
\small
You are given an image of a physics diagram.\\
You have generated the following description of the image with a confidence score:\\
{description}\\
\\
YOUR TASK is to read the question and select the correct answer from the provided options (shown in the image and described in your description) and provide a confidence score (0\% to 100\%) for your answer.\\
\\
During this time, thermal energy was transferred from () to ().\\
A. the surroundings . . . each salmon\\
B. each salmon . . . the surroundings\\
\\
\textbf{Instructions:}\\
- Carefully read and analyze the problem.\\
- Reason through the solution step by step, if helpful, considering both the image and your own description\\
- Take into account your confidence score of the description\\
- At the end, present your final answer and a confidence score in the following XML format:\\
<answer>[only the letter of the correct answer]</answer>\\
<confidence>[your confidence score here]</confidence>\\
\\
\textbf{Example output:}\\
\char91 YOUR\_REASONING\char93\\
<answer>A</answer>\\
<confidence>80\%</confidence>\\
\end{tcolorbox}

\subsection{Self-reflection prompting (1st round)}
Here, we present the prompts used in the first round of self-reflection.
\subsubsection{IsoBench}

\begin{tcolorbox}[remarkbox, title=graph\_maxflow\_image\_self\_reflection\_1st]
\small
You are given an image of a graph and two query nodes. (one source node and one sink node). \space \\
The source node is the node where the flow starts and the sink node is the node where the flow ends.\\
\\
YOUR TASK is to solve the maxflow problem given the weighted directed graph. \space \\
\\
\textbf{Definition of Maxflow problem:}\\
In the max flow problem, we have a directed graph with a source node s and a sink node t, and each edge has a capacity (integer valued, colored in green) that represents the maximum amount of flow that can be sent through it.\\
The goal is to find the maximum amount of flow that can be sent from s to t, while respecting the capacity constraints on the edges.\\
\\
\textbf{Query Example:}\\
Source node (zero-indexed): 0\\
Sink node (zero-indexed): 2\\
In the query example, the nodes are zero-indexed.\\
\\
\textbf{Instructions:}\\
- Please reason step by step\\
- At the end, present your final answer and a confidence score in the following XML format:\\
<answer>[the maximum flow from the source node to the sink node (in Arabic digits)]</answer>\\
\\
\textbf{Example output:}\\
\char91 YOUR\_REASONING\char93\\
<answer>12</answer>\\
\end{tcolorbox}

\begin{tcolorbox}[remarkbox, title=graph\_maxflow\_text\_self\_reflection\_1st]
\small
You are given an adjacency matrix of a graph and two query nodes. (one source node and one sink node). \space \\
The source node is the node where the flow starts and the sink node is the node where the flow ends.\\
\\
YOUR TASK is to solve the maxflow problem given the weighted directed graph. \space \\
\\
\textbf{Definition of Maxflow problem:}\\
In the max flow problem, we have a directed graph with a source node s and a sink node t, and each edge has a capacity that represents the maximum amount of flow that can be sent through it.\\
The goal is to find the maximum amount of flow that can be sent from s to t, while respecting the capacity constraints on the edges.\\
\\
\textbf{Query Example:}\\
adjacency matrix:\\
\char91 0, 1, 4\char93\\
\char91 0, 0, 6\char93\\
\char91 0, 0, 0\char93\\
Source node (zero-indexed): 0\\
Sink node (zero-indexed): 2\\
In the query example, the nodes are zero-indexed.\\
\\
\textbf{Instructions:}\\
- Please reason step by step\\
- At the end, present your final answer and a confidence score in the following XML format:\\
<answer>[the maximum flow from the source node to the sink node (in Arabic digits)]</answer>\\
\\
\textbf{Example output:}\\
\char91 YOUR\_REASONING\char93\\
<answer>12</answer>\\
\end{tcolorbox}

\begin{tcolorbox}[remarkbox, title=graph\_connectivity\_image\_self\_reflection\_1st]
\small
You are given an image of a graph and two query nodes. \space \\
\\
YOUR TASK is to determine whether the query nodes are connected as True or False.\\
\\
\textbf{Query Example:}\\
Query node 1 (zero-indexed): 9\\
Query node 2 (zero-indexed): 4\\
In the query example, the nodes are zero-indexed.\\
\\
\textbf{Instructions:}\\
- Please reason step by step\\
- At the end, present your final answer and a confidence score in the following XML format:\\
<answer>[whether the query nodes are connected: "True" or "False"]</answer>\\
\\
\textbf{Example output:}\\
\char91 YOUR\_REASONING\char93\\
<answer>True</answer>\\
\end{tcolorbox}

\begin{tcolorbox}[remarkbox, title=graph\_connectivity\_text\_self\_reflection\_1st]
\small
You are given the adjacency matrix of a graph and two query nodes. \space \\
\\
YOUR TASK is to determine whether the query nodes are connected as True or False.\\
\\
\textbf{Query Example:}\\
adjacency matrix:\\
\char91 0, 0, 0, 0, 0, 0, 0, 0, 0, 0, 0, 0\char93\\
\char91 0, 0, 0, 0, 0, 0, 0, 0, 0, 0, 0, 0\char93\\
\char91 0, 0, 0, 0, 0, 0, 0, 0, 0, 0, 0, 0\char93\\
\char91 0, 0, 0, 0, 0, 0, 0, 0, 0, 0, 0, 0\char93\\
\char91 0, 0, 0, 0, 0, 0, 0, 0, 0, 0, 0, 0\char93\\
\char91 0, 0, 0, 0, 0, 0, 0, 0, 0, 0, 0, 0\char93\\
\char91 0, 0, 0, 0, 0, 0, 0, 0, 0, 0, 0, 0\char93\\
\char91 0, 0, 0, 0, 0, 0, 0, 0, 0, 0, 0, 0\char93\\
\char91 0, 0, 0, 0, 0, 0, 0, 0, 0, 0, 1, 0\char93\\
\char91 0, 0, 0, 0, 0, 0, 0, 0, 0, 0, 0, 0\char93\\
\char91 0, 0, 0, 0, 0, 0, 0, 0, 1, 0, 0, 0\char93\\
\char91 0, 0, 0, 0, 0, 0, 0, 0, 0, 0, 0, 0\char93\\
Query node 1 (zero-indexed): 9\\
Query node 2 (zero-indexed): 4\\
In the query example, the nodes are zero-indexed.\\
\\
\textbf{Instructions:}\\
- Please reason step by step\\
- At the end, present your final answer and a confidence score in the following XML format:\\
<answer>[whether the query nodes are connected: "True" or "False"]</answer>\\
\\
\textbf{Example output:}\\
\char91 YOUR\_REASONING\char93\\
<answer>True</answer>\\
\end{tcolorbox}

\begin{tcolorbox}[remarkbox, title=graph\_isomorphism\_image\_self\_reflection\_1st]
\small
You are given an image of two specific graphs, G (Left Graph) and H (Right Graph).\\
\\
YOUR TASK is to determine if graph G and graph H are \textbf{isomorphic} based on the image.\\
\\
\textbf{Instructions:}\\
- Please reason step by step\\
- At the end, present your final answer and a confidence score in the following XML format:\\
<answer>[whether the two graphs are isomorphic: "True" or "False"]</answer>\\
\\
\textbf{Example output:}\\
\char91 YOUR\_REASONING\char93\\
<answer>True</answer>\\
\end{tcolorbox}

\begin{tcolorbox}[remarkbox, title=graph\_isomorphism\_text\_self\_reflection\_1st]
\small
You are given the adjacency matrix representations of two specific graphs, G and H.\\
\\
YOUR TASK is to determine if graph G and graph H, defined below, are \textbf{isomorphic} based on their provided adjacency matrices.\\
\\
\textbf{Query Example:}\\
adjacency matrix G:\\
\char91 0, 0, 0, 0, 0, 0, 0, 0, 0, 0, 0, 0\char93\\
\char91 0, 0, 0, 0, 0, 0, 0, 0, 0, 0, 0, 0\char93\\
\char91 0, 0, 0, 0, 0, 0, 0, 0, 0, 0, 0, 0\char93\\
\char91 0, 0, 0, 0, 0, 0, 0, 0, 0, 0, 0, 0\char93\\
\char91 0, 0, 0, 0, 0, 0, 0, 0, 0, 0, 0, 0\char93\\
\char91 0, 0, 0, 0, 0, 0, 0, 0, 0, 0, 0, 0\char93\\
\char91 0, 0, 0, 0, 0, 0, 0, 0, 0, 0, 0, 0\char93\\
\char91 0, 0, 0, 0, 0, 0, 0, 0, 0, 0, 0, 0\char93\\
\char91 0, 0, 0, 0, 0, 0, 0, 0, 0, 0, 1, 0\char93\\
\char91 0, 0, 0, 0, 0, 0, 0, 0, 0, 0, 0, 0\char93\\
\char91 0, 0, 0, 0, 0, 0, 0, 0, 1, 0, 0, 0\char93\\
\char91 0, 0, 0, 0, 0, 0, 0, 0, 0, 0, 0, 0\char93\\
adjacency matrix H:\\
\char91 0, 0, 0, 0, 0, 0, 0, 0, 0, 0, 0, 0\char93\\
\char91 0, 0, 0, 0, 0, 0, 0, 0, 0, 0, 0, 0\char93\\
\char91 0, 0, 0, 0, 0, 0, 0, 0, 0, 0, 0, 0\char93\\
\char91 0, 0, 0, 0, 0, 0, 0, 0, 0, 0, 0, 0\char93\\
\char91 0, 0, 0, 0, 0, 0, 0, 0, 0, 0, 0, 0\char93\\
\char91 0, 0, 0, 0, 0, 0, 0, 0, 0, 0, 0, 0\char93\\
\char91 0, 0, 0, 0, 0, 0, 0, 0, 0, 0, 0, 0\char93\\
\char91 0, 0, 0, 0, 0, 0, 0, 0, 0, 1, 0, 0\char93\\
\char91 0, 0, 0, 0, 0, 0, 0, 0, 0, 0, 0, 0\char93\\
\char91 0, 0, 0, 0, 0, 0, 0, 1, 0, 0, 0, 0\char93\\
\char91 0, 0, 0, 0, 0, 0, 0, 0, 0, 0, 0, 0\char93\\
\char91 0, 0, 0, 0, 0, 0, 0, 0, 0, 0, 0, 0\char93\\
\\
\textbf{Instructions:}\\
- Please reason step by step\\
- At the end, present your final answer and a confidence score in the following XML format:\\
<answer>[whether the two graphs are isomorphic: "True" or "False"]</answer>\\
\\
\textbf{Example output:}\\
\char91 YOUR\_REASONING\char93\\
<answer>True</answer>\\
\end{tcolorbox}

\begin{tcolorbox}[remarkbox, title=puzzle\_image\_self\_reflection\_1st]
\small
You are given a visual representation of a chess puzzle for which a sequence of unique best moves is determinable (e.g. sequences of moves leading to a forced checkmate).\\
\\
\textbf{Definition of the Chess Puzzle:}\\
In a chess puzzle, you are required to make a series of optimal moves leading to checkmate, starting from the given position.\\
\\
YOUR TASK is to predict THE FIRST MOVE that should be played given this board setup.\\
Your answer should specify the move in Algebraic Coordinate Notation (e.g., "d2d1", "e5a1", "c4f4").\\
\\
\textbf{Instructions:}\\
- Please reason step by step\\
- At the end, present your final answer and a confidence score in the following XML format:\\
<answer>[only the first move in Algebraic Coordinate Notation]</answer>\\
\\
\textbf{Example output:}\\
\char91 YOUR\_REASONING\char93\\
<answer>e2e4</answer>\\
\end{tcolorbox}

\begin{tcolorbox}[remarkbox, title=puzzle\_pgn\_self\_reflection\_1st]
\small
You are given a PGN representation of a chess puzzle for which a sequence of unique best moves is determinable (e.g. sequences of moves leading to a forced checkmate).\\
\\
\textbf{Definition of the Chess Puzzle:}\\
In a chess puzzle, you are required to make a series of optimal moves leading to checkmate, starting from the given position.\\
\\
YOUR TASK is to predict THE FIRST MOVE that should be played given this board setup.\\
Your answer should specify the move in Algebraic Coordinate Notation (e.g., "d2d1", "e5a1", "c4f4").\\
\\
PGN: 1. e4 e6 2. d4 Ne7 3. c4 Ng6 4. Nf3 Nh4 5. Nxh4 Qxh4 6. Bd3 b6 7. O-O Bb7 8. Nc3 Nc6 9. d5 Ne7 10. Qf3 Ng6 11. Qg3 Qxg3 12. fxg3 Ne5 13. Be2 Bc5+ 14. Kh1 O-O 15. Bf4 Bd4 16. Rad1 Bxc3 17. bxc3 Ng6 18. Bxc7 exd5 19. cxd5 Rfe8 20. Bf3 Ne5 21. Bxe5 Rxe5 22. c4 Ba6 23. Rc1 d6 24. Rfe1 Rae8 25. Kg1 R8e7 26. Kf2 f5 27. exf5 Rxe1 28. Rxe1 Rxe1 29. Kxe1 Bxc4 30. a3 a5 31. Kd2 Kf7 32. Kc3 Bf1 33. h4 Kf6 34. g4 Ke5 35. h5 h6 36. Kb3 Kd4 37. Ka4 Bc4 38. g3 Ba6 39. g5 hxg5 40. f6 gxf6 41. h6 Bd3 42. g4 Kc5 43. Be2 Bh7 44. Bb5 Kxd5 45. Bd7 Bg8 46. Bf5 Ke5 47. h7 Bxh7 48. Bxh7 d5 49. Kb5 d4 50. Kc4 a4 51. Bc2 b5+ 52. Kxb5 Kf4 53. Bd1 d3 54. Kxa4 f5\\
\\
\textbf{Instructions:}\\
- Please reason step by step\\
- At the end, present your final answer and a confidence score in the following XML format:\\
<answer>[only the first move in Algebraic Coordinate Notation]</answer>\\
\\
\textbf{Example output:}\\
\char91 YOUR\_REASONING\char93\\
<answer>e2e4</answer>\\
\end{tcolorbox}

\begin{tcolorbox}[remarkbox, title=winner\_id\_image\_self\_reflection\_1st]
\small
You are given a visual representation of a chess puzzle for which a sequence of unique best moves is determinable (e.g. sequences of moves leading to a forced checkmate).\\
\\
\textbf{Definition of the Chess Puzzle:}\\
In a chess puzzle, you are required to make a series of optimal moves leading to checkmate, starting from the given position.\\
\\
YOUR TASK is to identify the winner of this game given this board setup.\\
Your answer should specify the winner as one of the following strings: "White", "Black", or "Draw".\\
\\
\textbf{Instructions:}\\
- Please reason step by step\\
- At the end, present your final answer and a confidence score in the following XML format:\\
<answer>[only the winner of this game: "White", "Black", or "Draw"]</answer>\\
\\
\textbf{Example output:}\\
\char91 YOUR\_REASONING\char93\\
<answer>Draw</answer>\\
\end{tcolorbox}

\begin{tcolorbox}[remarkbox, title=winner\_id\_pgn\_self\_reflection\_1st]
\small
You are given a PGN representation of a chess puzzle for which a sequence of unique best moves is determinable (e.g. sequences of moves leading to a forced checkmate).\\
\\
\textbf{Definition of the Chess Puzzle:}\\
In a chess puzzle, you are required to make a series of optimal moves leading to checkmate, starting from the given position.\\
\\
YOUR TASK is to identify the winner of this game given this board setup.\\
Your answer should specify the winner as one of the following strings: "White", "Black", or "Draw".\\
\\
PGN: 1. d4 d5 2. e3 e6 3. Bd3 Nf6 4. Nd2 Be7 5. c3 O-O 6. f4 Nbd7 7. Qe2 c5 8. Ngf3 c4 9. Bc2 a6 10. O-O b5 11. Ne5 Bb7 12. a3 Rb8 13. e4 dxe4 14. Nxe4 Nxe5 15. fxe5 Nd5 16. Qg4 a5 17. Bh6 f6 18. Qxg7\#\\
\\
\textbf{Instructions:}\\
- Please reason step by step\\
- At the end, present your final answer and a confidence score in the following XML format:\\
<answer>[only the winner of this game: "White", "Black", or "Draw"]</answer>\\
\\
\textbf{Example output:}\\
\char91 YOUR\_REASONING\char93\\
<answer>Draw</answer>\\
\end{tcolorbox}

\begin{tcolorbox}[remarkbox, title=image\_math\_parity\_self\_reflection\_1st]
\small
You are given a plot of a real-valued, scalar function f(x).\\
YOUR TASK is to determine whether f(x) is an even function, an odd function, or neither.\\
- Definition of an even function: A function such that f(x) = f(-x) where the value remains unchanged if the sign of the independent variable is reversed.\\
- Definition of an odd function: A function such that f(-x) = -f(x) where the sign is reversed but the absolute value remains the same if the sign of the independent variable is reversed\\
- A function is neither even nor odd if it does not satisfy either definitions.\\
\\
\textbf{Instructions:}\\
- Please reason step by step\\
- At the end, present your final answer in the following XML format:\\
<answer>[only the final result: 'even', 'odd', or 'neither']</answer>\\
\\
\textbf{Example output:}\\
\char91 YOUR\_REASONING\char93\\
<answer>even</answer>\\
\end{tcolorbox}

\begin{tcolorbox}[remarkbox, title=image\_math\_convexity\_self\_reflection\_1st]
\small
You are given a plot of a real-valued, scalar function f(x).\\
YOUR TASK is to determine whether f(x) is an convex function or a concave function\\
- Definition of a convex function: A function such that for all x, y, and $0 \leq t \leq 1$\\
$f(tx + (1 - t)y) \leq tf(x) + (1 - t)f(y)$\\
- Definition of a concave function: A function such that for all x, y, and $0 \leq t \leq 1$\\
$f(tx + (1 - t)y) \geq tf(x) + (1 - t)f(y)$\\
\\
\textbf{Instructions:}\\
- Please reason step by step\\
- At the end, present your final answer in the following XML format:\\
<answer>[only the final result: 'convex' or 'concave']</answer>\\
\\
\textbf{Example output:}\\
\char91 YOUR\_REASONING\char93\\
<answer>convex</answer>\\
\end{tcolorbox}

\begin{tcolorbox}[remarkbox, title=image\_math\_breakpoint\_self\_reflection\_1st]
\small
You are given a plot of a real-valued, scalar function f(x).\\
YOUR TASK is to count the number of breakpoints in the plot of f(x). \space \\
A breakpoint refers to a point on the function’s domain at which the function changes its slope.\\
\\
You should IGNORE the left and right end point of the domain, i.e. if the function is defined on [a, b], you should only consider the domain (a, b).\\
\\
\textbf{Instructions:}\\
- Please reason step by step\\
- At the end, present your final answer in the following XML format:\\
<answer>[the number of breakpoints (in Arabic digits)]</answer>\\
\\
\textbf{Example output:}\\
\char91 YOUR\_REASONING\char93\\
<answer>2</answer>\\
\end{tcolorbox}

\begin{tcolorbox}[remarkbox, title=text\_math\_parity\_self\_reflection\_1st]
\small
You are given a real-valued, scalar function f(x).\\
YOUR TASK is to determine whether f(x) is an even function, an odd function, or neither.\\
- Definition of an even function: A function such that f(x) = f(-x) where the value remains unchanged if the sign of the independent variable is reversed.\\
- Definition of an odd function: A function such that f(-x) = -f(x) where the sign is reversed but the absolute value remains the same if the sign of the independent variable is reversed\\
- A function is neither even nor odd if it does not satisfy either definitions.\\
\\
Here is the expression of f(x){domain}:\\
\{text\}\\
\\
\textbf{Instructions:}\\
- Please reason step by step\\
- At the end, present your final answer in the following XML format:\\
<answer>[only the final result: 'even', 'odd', or 'neither']</answer>\\
\\
\textbf{Example output:}\\
\char91 YOUR\_REASONING\char93\\
<answer>even</answer>\\
\end{tcolorbox}

\begin{tcolorbox}[remarkbox, title=text\_math\_convexity\_self\_reflection\_1st]
\small
You are given a real-valued, scalar function f(x).\\
YOUR TASK is to determine whether f(x) is an convex function or a concave function\\
- Definition of a convex function: A function such that for all x, y, and $0 \leq t \leq 1$\\
$f(tx + (1 - t)y) \leq tf(x) + (1 - t)f(y)$\\
- Definition of a concave function: A function such that for all x, y, and $0 \leq t \leq 1$\\
$f(tx + (1 - t)y) \geq tf(x) + (1 - t)f(y)$\\
\\
Here is the expression of f(x){domain}:\\
\{text\}\\
\\
\textbf{Instructions:}\\
- Please reason step by step\\
- At the end, present your final answer in the following XML format:\\
<answer>[only the final result: 'convex' or 'concave']</answer>\\
\\
\textbf{Example output:}\\
\char91 YOUR\_REASONING\char93\\
<answer>convex</answer>\\
\end{tcolorbox}

\begin{tcolorbox}[remarkbox, title=text\_math\_breakpoint\_self\_reflection\_1st]
\small
You are given a real-valued, scalar function f(x).\\
YOUR TASK is to count the number of breakpoints in the plot of f(x). \space \\
A breakpoint refers to a point on the function’s domain at which the function changes its slope.\\
\\
Here is the expression of f(x){domain}:\\
\{text\}\\
\\
You should IGNORE the left and right end point of the domain, i.e. if the function is defined on [a, b], you should only consider the domain (a, b).\\
\\
\textbf{Instructions:}\\
- Please reason step by step\\
- At the end, present your final answer in the following XML format:\\
<answer>[the number of breakpoints (in Arabic digits)]</answer>\\
\\
\textbf{Example output:}\\
\char91 YOUR\_REASONING\char93\\
<answer>2</answer>\\
\end{tcolorbox}

\begin{tcolorbox}[remarkbox, title=chemistry\_image\_self\_reflection\_1st]
\small
You are given an image of a chemistry diagram.\\
YOUR TASK is to read the question and select the correct answer from the provided options.\\
\\
Which solution has a higher concentration of green particles?\\
A. Solution B\\
B. neither; their concentrations are the same\\
C. Solution A\\
\\
\textbf{Instructions:}\\
- Carefully read and analyze the problem.\\
- Reason through the solution step by step, if helpful.\\
- At the end, present your final answer in the following XML format:\\
<answer>[only the letter of the correct answer]</answer>\\
\\
\textbf{Example output:}\\
\char91 YOUR\_REASONING\char93\\
<answer>A</answer>\\
\end{tcolorbox}

\begin{tcolorbox}[remarkbox, title=chemistry\_text\_self\_reflection\_1st]
\small
You are given a multiple-choice chemistry question.\\
YOUR TASK is to read the question and select the correct answer from the provided options.\\
\\
In Solution A and Solution B, the green particles represent the solute. The volume of the solvent in two containers are equal. Solution A and Solution B have the same number of green particles.\\
\\
Which solution has a higher concentration of green particles?\\
A. Solution B\\
B. neither; their concentrations are the same\\
C. Solution A\\
\\
\textbf{Instructions:}\\
- Carefully read and analyze the problem.\\
- Reason through the solution step by step, if helpful.\\
- At the end, present your final answer in the following XML format:\\
<answer>[only the letter of the correct answer]</answer>\\
\\
\textbf{Example output:}\\
\char91 YOUR\_REASONING\char93\\
<answer>A</answer>\\
\end{tcolorbox}

\begin{tcolorbox}[remarkbox, title=physics\_image\_self\_reflection\_1st]
\small
You are given an image of a physics diagram.\\
YOUR TASK is to read the question and select the correct answer from the provided options.\\
\\
During this time, thermal energy was transferred from () to ().\\
A. the surroundings . . . each salmon\\
B. each salmon . . . the surroundings\\
\\
\textbf{Instructions:}\\
- Carefully read and analyze the problem.\\
- Reason through the solution step by step, if helpful.\\
- At the end, present your final answer in the following XML format:\\
<answer>[only the letter of the correct answer]</answer>\\
\\
\textbf{Example output:}\\
\char91 YOUR\_REASONING\char93\\
<answer>A</answer>\\
\end{tcolorbox}

\begin{tcolorbox}[remarkbox, title=physics\_text\_self\_reflection\_1st]
\small
You are given a multiple-choice physics question.\\
YOUR TASK is to read the question and select the correct answer from the provided options.\\
\\
The temperature of each salmon increased.\\
\\
During this time, thermal energy was transferred from () to ().\\
A. the surroundings . . . each salmon\\
B. each salmon . . . the surroundings\\
\\
\textbf{Instructions:}\\
- Carefully read and analyze the problem.\\
- Reason through the solution step by step, if helpful.\\
- At the end, present your final answer in the following XML format:\\
<answer>[only the letter of the correct answer]</answer>\\
\\
\textbf{Example output:}\\
\char91 YOUR\_REASONING\char93\\
<answer>A</answer>\\
\end{tcolorbox}

\subsection{Self-reflection prompting (2nd round)}
We use a shared prompt template for the second round of all self-reflection experiments:
\begin{tcolorbox}[remarkbox, title=shared\_self\_reflection\_2nd]
\small
\textbf{Task:} Reflect on the following problem and solution, and provide a final confidence score to the solution.
\\\\
\textbf{Instructions:}\\
- Carefully read and analyze the problem and solution.\\
- Reason through the solution step by step, if helpful.\\
- At the end, present your final answer in the following XML format:\\
<confidence>confidence score here</confidence>\\
\\
\textbf{Example output:} \\
\char91 YOUR\_REASONING\char93\\
<confidence>80\%</confidence>\\
\\
\textbf{Now, here is the problem and solution:} \\
\textbf{Problem:}\\
\{problem\}\\\\
\textbf{Solution:}\\
\{solution\}
\end{tcolorbox}

\subsection{Top-K prompting}
Here, we present the prompts used in the Top-K prompting. In our experiments, we use K=3.

\subsubsection{IsoBench}

\begin{tcolorbox}[remarkbox, title=graph\_maxflow\_image\_topk]
\small
You are given an image of a graph and two query nodes. (one source node and one sink node). \space \\
The source node is the node where the flow starts and the sink node is the node where the flow ends.\\
\\
YOUR TASK is to solve the maxflow problem given the weighted directed graph.\\
\\
\textbf{Definition of Maxflow problem:}\\
In the max flow problem, we have a directed graph with a source node s and a sink node t, and each edge has a capacity (integer valued, colored in green) that represents the maximum amount of flow that can be sent through it.\\
The goal is to find the maximum amount of flow that can be sent from s to t, while respecting the capacity constraints on the edges.\\
\\
\textbf{Query Example:}\\
Source node (zero-indexed): 0\\
Sink node (zero-indexed): 2\\
In the query example, the nodes are zero-indexed.\\
\\
\textbf{Instructions:}\\
- Provide your 3 best guesses and the probability that each is correct (0\% to 100\%) for the following question.\\
- If the number of options for tge question is less than 3, only provide the confidence score for each option as the answer.\\
- Please reason step by step\\
- At the end, present your final answers and confidence scores in the following XML format:\\
<answer1>[the maximum flow from the source node to the sink node (in Arabic digits)]</answer1>\\
<confidence1>[your confidence score for the answer1]</confidence1>\\
<answer2>[the maximum flow from the source node to the sink node (in Arabic digits)]</answer2>\\
<confidence2>[your confidence score for the answer2]</confidence2>\\
<answer3>[the maximum flow from the source node to the sink node (in Arabic digits)]</answer3>\\
<confidence3>[your confidence score for the answer3]</confidence3>\\
\\
\textbf{Example output:}\\
\char91 YOUR\_REASONING\char93\\
<answer1>16</answer1>\\
<confidence1>95\%</confidence1>\\
<answer2>12</answer2>\\
<confidence2>80\%</confidence2>\\
<answer3>23</answer3>\\
<confidence3>50\%</confidence3>\\
\end{tcolorbox}

\begin{tcolorbox}[remarkbox, title=graph\_maxflow\_text\_topk]
\small
You are given an adjacency matrix of a graph and two query nodes. (one source node and one sink node). \space \\
The source node is the node where the flow starts and the sink node is the node where the flow ends.\\
\\
YOUR TASK is to solve the maxflow problem given the weighted directed graph.\\
\\
\textbf{Definition of Maxflow problem:}\\
In the max flow problem, we have a directed graph with a source node s and a sink node t, and each edge has a capacity that represents the maximum amount of flow that can be sent through it.\\
The goal is to find the maximum amount of flow that can be sent from s to t, while respecting the capacity constraints on the edges.\\
\\
\textbf{Query Example:}\\
adjacency matrix:\\
\char91 0, 1, 4\char93\\
\char91 0, 0, 6\char93\\
\char91 0, 0, 0\char93\\
Source node (zero-indexed): 0\\
Sink node (zero-indexed): 2\\
In the query example, the nodes are zero-indexed.\\
\\
\textbf{Instructions:}\\
- Provide your 3 best guesses and the probability that each is correct (0\% to 100\%) for the following question.\\
- If the number of options for the question is less than 3, only provide the confidence score for each option as the answer.\\
- Please reason step by step\\
- At the end, present your final answers and confidence scores in the following XML format:\\
<answer1>[the maximum flow from the source node to the sink node (in Arabic digits)]</answer1>\\
<confidence1>[your confidence score for the answer1]</confidence1>\\
<answer2>[the maximum flow from the source node to the sink node (in Arabic digits)]</answer2>\\
<confidence2>[your confidence score for the answer2]</confidence2>\\
<answer3>[the maximum flow from the source node to the sink node (in Arabic digits)]</answer3>\\
<confidence3>[your confidence score for the answer3]</confidence3>\\
\\
\textbf{Example output:}\\
\char91 YOUR\_REASONING\char93\\
<answer1>16</answer1>\\
<confidence1>95\%</confidence1>\\
<answer2>12</answer2>\\
<confidence2>80\%</confidence2>\\
<answer3>23</answer3>\\
<confidence3>50\%</confidence3>\\
\end{tcolorbox}

\begin{tcolorbox}[remarkbox, title=graph\_connectivity\_image\_topk]
\small
You are given an image of a graph and two query nodes. \space \\
\\
YOUR TASK is to determine whether the query nodes are connected as True or False.\\
\\
\textbf{Query Example:}\\
Query node 1 (zero-indexed): 9\\
Query node 2 (zero-indexed): 4\\
In the query example, the nodes are zero-indexed.\\
\\
\textbf{Instructions:}\\
- Provide your 3 best guesses and the probability that each is correct (0\% to 100\%) for the following question.\\
- If the number of options for the question is less than 3, only provide the confidence score for each option as the answer.\\
- Please reason step by step\\
- At the end, present your final answers and confidence scores in the following XML format:\\
<answer1>[whether the query nodes are connected: "True" or "False"]</answer1>\\
<confidence1>[your confidence score for the answer1]</confidence1>\\
<answer2>[whether the query nodes are connected: "True" or "False"]</answer2>\\
<confidence2>[your confidence score for the answer2]</confidence2>\\
\\
\textbf{Example output:}\\
\char91 YOUR\_REASONING\char93\\
<answer1>True</answer1>\\
<confidence1>80\%</confidence1>\\
<answer2>False</answer2>\\
<confidence2>60\%</confidence2>\\
\end{tcolorbox}

\begin{tcolorbox}[remarkbox, title=graph\_connectivity\_text\_topk]
\small
You are given the adjacency matrix of a graph and two query nodes. \space \\
\\
YOUR TASK is to determine whether the query nodes are connected as True or False.\\
\\
\textbf{Query Example:}\\
adjacency matrix:\\
\char91 0, 0, 0, 0, 0, 0, 0, 0, 0, 0, 0, 0\char93\\
\char91 0, 0, 0, 0, 0, 0, 0, 0, 0, 0, 0, 0\char93\\
\char91 0, 0, 0, 0, 0, 0, 0, 0, 0, 0, 0, 0\char93\\
\char91 0, 0, 0, 0, 0, 0, 0, 0, 0, 0, 0, 0\char93\\
\char91 0, 0, 0, 0, 0, 0, 0, 0, 0, 0, 0, 0\char93\\
\char91 0, 0, 0, 0, 0, 0, 0, 0, 0, 0, 0, 0\char93\\
\char91 0, 0, 0, 0, 0, 0, 0, 0, 0, 0, 0, 0\char93\\
\char91 0, 0, 0, 0, 0, 0, 0, 0, 0, 0, 0, 0\char93\\
\char91 0, 0, 0, 0, 0, 0, 0, 0, 0, 0, 1, 0\char93\\
\char91 0, 0, 0, 0, 0, 0, 0, 0, 0, 0, 0, 0\char93\\
\char91 0, 0, 0, 0, 0, 0, 0, 0, 1, 0, 0, 0\char93\\
\char91 0, 0, 0, 0, 0, 0, 0, 0, 0, 0, 0, 0\char93\\
Query node 1 (zero-indexed): 9\\
Query node 2 (zero-indexed): 4\\
In the query example, the nodes are zero-indexed.\\
\\
\textbf{Instructions:}\\
- Provide your 3 best guesses and the probability that each is correct (0\% to 100\%) for the following question.\\
- If the number of options for the question is less than 3, only provide the confidence score for each option as the answer.\\
- Please reason step by step\\
- At the end, present your final answers and confidence scores in the following XML format:\\
<answer1>[whether the query nodes are connected: "True" or "False"]</answer1>\\
<confidence1>[your confidence score for the answer1]</confidence1>\\
<answer2>[whether the query nodes are connected: "True" or "False"]</answer2>\\
<confidence2>[your confidence score for the answer2]</confidence2>\\
\\
\textbf{Example output:}\\
\char91 YOUR\_REASONING\char93\\
<answer1>True</answer1>\\
<confidence1>80\%</confidence1>\\
<answer2>False</answer2>\\
<confidence2>60\%</confidence2>\\
\end{tcolorbox}

\begin{tcolorbox}[remarkbox, title=graph\_isomorphism\_image\_topk]
\small
You are given an image of two specific graphs, G (Left Graph) and H (Right Graph).\\
\\
YOUR TASK is to determine if graph G and graph H are \textbf{isomorphic} based on the image, and provide a confidence score (0\% to 100\%) for your determination.\\
\\
\textbf{Instructions:}\\
- Provide your 3 best guesses and the probability that each is correct (0\% to 100\%) for the following question.\\
- If the number of options for the question is less than 3, only provide the confidence score for each option as the answer.\\
- Please reason step by step\\
- At the end, present your final answers and confidence scores in the following XML format:\\
<answer1>[whether the two graphs are isomorphic: "True" or "False"]</answer1>\\
<confidence1>[your confidence score for the answer1]</confidence1>\\
<answer2>[whether the two graphs are isomorphic: "True" or "False"]</answer2>\\
<confidence2>[your confidence score for the answer2]</confidence2>\\
\\
\textbf{Example output:}\\
\char91 YOUR\_REASONING\char93\\
<answer1>True</answer1>\\
<confidence1>80\%</confidence1>\\
<answer2>False</answer2>\\
<confidence2>60\%</confidence2>\\
\end{tcolorbox}

\begin{tcolorbox}[remarkbox, title=graph\_isomorphism\_text\_topk]
\small
You are given the adjacency matrix representations of two specific graphs, G and H.\\
\\
YOUR TASK is to determine if graph G and graph H, defined below, are \textbf{isomorphic} based on their provided adjacency matrices.\\
\\
\textbf{Query Example:}\\
adjacency matrix G:\\
\char91 0, 0, 0, 0, 0, 0, 0, 0, 0, 0, 0, 0\char93\\
\char91 0, 0, 0, 0, 0, 0, 0, 0, 0, 0, 0, 0\char93\\
\char91 0, 0, 0, 0, 0, 0, 0, 0, 0, 0, 0, 0\char93\\
\char91 0, 0, 0, 0, 0, 0, 0, 0, 0, 0, 0, 0\char93\\
\char91 0, 0, 0, 0, 0, 0, 0, 0, 0, 0, 0, 0\char93\\
\char91 0, 0, 0, 0, 0, 0, 0, 0, 0, 0, 0, 0\char93\\
\char91 0, 0, 0, 0, 0, 0, 0, 0, 0, 0, 0, 0\char93\\
\char91 0, 0, 0, 0, 0, 0, 0, 0, 0, 0, 0, 0\char93\\
\char91 0, 0, 0, 0, 0, 0, 0, 0, 0, 0, 1, 0\char93\\
\char91 0, 0, 0, 0, 0, 0, 0, 0, 0, 0, 0, 0\char93\\
\char91 0, 0, 0, 0, 0, 0, 0, 0, 1, 0, 0, 0\char93\\
\char91 0, 0, 0, 0, 0, 0, 0, 0, 0, 0, 0, 0\char93\\
adjacency matrix H:\\
\char91 0, 0, 0, 0, 0, 0, 0, 0, 0, 0, 0, 0\char93\\
\char91 0, 0, 0, 0, 0, 0, 0, 0, 0, 0, 0, 0\char93\\
\char91 0, 0, 0, 0, 0, 0, 0, 0, 0, 0, 0, 0\char93\\
\char91 0, 0, 0, 0, 0, 0, 0, 0, 0, 0, 0, 0\char93\\
\char91 0, 0, 0, 0, 0, 0, 0, 0, 0, 0, 0, 0\char93\\
\char91 0, 0, 0, 0, 0, 0, 0, 0, 0, 0, 0, 0\char93\\
\char91 0, 0, 0, 0, 0, 0, 0, 0, 0, 0, 0, 0\char93\\
\char91 0, 0, 0, 0, 0, 0, 0, 0, 0, 1, 0, 0\char93\\
\char91 0, 0, 0, 0, 0, 0, 0, 0, 0, 0, 0, 0\char93\\
\char91 0, 0, 0, 0, 0, 0, 0, 1, 0, 0, 0, 0\char93\\
\char91 0, 0, 0, 0, 0, 0, 0, 0, 0, 0, 0, 0\char93\\
\char91 0, 0, 0, 0, 0, 0, 0, 0, 0, 0, 0, 0\char93\\
\\
\textbf{Instructions:}\\
- Provide your 3 best guesses and the probability that each is correct (0\% to 100\%) for the following question.\\
- If the number of options for the question is less than 3, only provide the confidence score for each option as the answer.\\
- Please reason step by step\\
- At the end, present your final answers and confidence scores in the following XML format:\\
<answer1>[whether the two graphs are isomorphic: "True" or "False"]</answer1>\\
<confidence1>[your confidence score for the answer1]</confidence1>\\
<answer2>[whether the two graphs are isomorphic: "True" or "False"]</answer2>\\
<confidence2>[your confidence score for the answer2]</confidence2>\\
\\
\textbf{Example output:}\\
\char91 YOUR\_REASONING\char93\\
<answer1>True</answer1>\\
<confidence1>80\%</confidence1>\\
<answer2>False</answer2>\\
<confidence2>60\%</confidence2>\\
\end{tcolorbox}

\begin{tcolorbox}[remarkbox, title=puzzle\_image\_topk]
\small
You are given a visual representation of a chess puzzle for which a sequence of unique best moves is determinable (e.g. sequences of moves leading to a forced checkmate).\\
\\
\textbf{Definition of the Chess Puzzle:}\\
In a chess puzzle, you are required to make a series of optimal moves leading to checkmate, starting from the given position.\\
\\
YOUR TASK is to predict THE FIRST MOVE that should be played given this board setup.\\
Your answer should specify the move in Algebraic Coordinate Notation (e.g., "d2d1", "e5a1", "c4f4").\\
\\
\textbf{Instructions:}\\
- Provide your 3 best guesses and the probability that each is correct (0\% to 100\%) for the following question.\\
- If the number of options for the question is less than 3, only provide the confidence score for each option as the answer.\\
- Please reason step by step\\
- At the end, present your final answers and confidence scores in the following XML format:\\
<answer1>[only the first move in Algebraic Coordinate Notation]</answer1>\\
<confidence1>[your confidence score for the answer1]</confidence1>\\
<answer2>[only the first move in Algebraic Coordinate Notation]</answer2>\\
<confidence2>[your confidence score for the answer2]</confidence2>\\
<answer3>[only the first move in Algebraic Coordinate Notation]</answer3>\\
<confidence3>[your confidence score for the answer3]</confidence3>\\
\\
\textbf{Example output:}\\
\char91 YOUR\_REASONING\char93\\
<answer1>e2e4</answer1>\\
<confidence1>95\%</confidence1>\\
<answer2>e1f2</answer2>\\
<confidence2>80\%</confidence2>\\
<answer3>d2a3</answer3>\\
<confidence3>50\%</confidence3>\\
\end{tcolorbox}

\begin{tcolorbox}[remarkbox, title=puzzle\_pgn\_topk]
\small
You are given a PGN representation of a chess puzzle for which a sequence of unique best moves is determinable (e.g. sequences of moves leading to a forced checkmate).\\
\\
\textbf{Definition of the Chess Puzzle:}\\
In a chess puzzle, you are required to make a series of optimal moves leading to checkmate, starting from the given position.\\
\\
YOUR TASK is to predict THE FIRST MOVE that should be played given this board setup.\\
Your answer should specify the move in Algebraic Coordinate Notation (e.g., "d2d1", "e5a1", "c4f4").\\
\\
PGN: 1. e4 e6 2. d4 Ne7 3. c4 Ng6 4. Nf3 Nh4 5. Nxh4 Qxh4 6. Bd3 b6 7. O-O Bb7 8. Nc3 Nc6 9. d5 Ne7 10. Qf3 Ng6 11. Qg3 Qxg3 12. fxg3 Ne5 13. Be2 Bc5+ 14. Kh1 O-O 15. Bf4 Bd4 16. Rad1 Bxc3 17. bxc3 Ng6 18. Bxc7 exd5 19. cxd5 Rfe8 20. Bf3 Ne5 21. Bxe5 Rxe5 22. c4 Ba6 23. Rc1 d6 24. Rfe1 Rae8 25. Kg1 R8e7 26. Kf2 f5 27. exf5 Rxe1 28. Rxe1 Rxe1 29. Kxe1 Bxc4 30. a3 a5 31. Kd2 Kf7 32. Kc3 Bf1 33. h4 Kf6 34. g4 Ke5 35. h5 h6 36. Kb3 Kd4 37. Ka4 Bc4 38. g3 Ba6 39. g5 hxg5 40. f6 gxf6 41. h6 Bd3 42. g4 Kc5 43. Be2 Bh7 44. Bb5 Kxd5 45. Bd7 Bg8 46. Bf5 Ke5 47. h7 Bxh7 48. Bxh7 d5 49. Kb5 d4 50. Kc4 a4 51. Bc2 b5+ 52. Kxb5 Kf4 53. Bd1 d3 54. Kxa4 f5\\
\\
\textbf{Instructions:}\\
- Provide your 3 best guesses and the probability that each is correct (0\% to 100\%) for the following question.\\
- If the number of options for the question is less than 3, only provide the confidence score for each option as the answer.\\
- Please reason step by step\\
- At the end, present your final answers and confidence scores in the following XML format:\\
<answer1>[only the first move in Algebraic Coordinate Notation]</answer1>\\
<confidence1>[your confidence score for the answer1]</confidence1>\\
<answer2>[only the first move in Algebraic Coordinate Notation]</answer2>\\
<confidence2>[your confidence score for the answer2]</confidence2>\\
<answer3>[only the first move in Algebraic Coordinate Notation]</answer3>\\
<confidence3>[your confidence score for the answer3]</confidence3>\\
\\
\textbf{Example output:}\\
\char91 YOUR\_REASONING\char93\\
<answer1>e2e4</answer1>\\
<confidence1>95\%</confidence1>\\
<answer2>e1f2</answer2>\\
<confidence2>80\%</confidence2>\\
<answer3>d2a3</answer3>\\
<confidence3>50\%</confidence3>\\
\end{tcolorbox}

\begin{tcolorbox}[remarkbox, title=winner\_id\_image\_topk]
\small
You are given a visual representation of a chess puzzle for which a sequence of unique best moves is determinable (e.g. sequences of moves leading to a forced checkmate).\\
\\
\textbf{Definition of the Chess Puzzle:}\\
In a chess puzzle, you are required to make a series of optimal moves leading to checkmate, starting from the given position.\\
\\
YOUR TASK is to identify the winner of this game given this board setup.\\
Your answer should specify the winner as one of the following strings: "White", "Black", or "Draw".\\
\\
\textbf{Instructions:}\\
- Provide your 3 best guesses and the probability that each is correct (0\% to 100\%) for the following question.\\
- If the number of options for the question is less than 3, only provide the confidence score for each option as the answer.\\
- Please reason step by step\\
- At the end, present your final answers and confidence scores in the following XML format:\\
<answer1>[only the winner of this game: "White", "Black", or "Draw"]</answer1>\\
<confidence1>[your confidence score for the answer1]</confidence1>\\
<answer2>[only the winner of this game: "White", "Black", or "Draw"]</answer2>\\
<confidence2>[your confidence score for the answer2]</confidence2>\\
<answer3>[only the winner of this game: "White", "Black", or "Draw"]</answer3>\\
<confidence3>[your confidence score for the answer3]</confidence3>\\
\\
\textbf{Example output:}\\
\char91 YOUR\_REASONING\char93\\
<answer1>Draw</answer1>\\
<confidence1>90\%</confidence1>\\
<answer2>Black</answer2>\\
<confidence2>80\%</confidence2>\\
<answer3>White</answer3>\\
<confidence3>50\%</confidence3>\\
\end{tcolorbox}

\begin{tcolorbox}[remarkbox, title=winner\_id\_pgn\_topk]
\small
You are given a PGN representation of a chess puzzle for which a sequence of unique best moves is determinable (e.g. sequences of moves leading to a forced checkmate).\\
\\
\textbf{Definition of the Chess Puzzle:}\\
In a chess puzzle, you are required to make a series of optimal moves leading to checkmate, starting from the given position.\\
\\
YOUR TASK is to identify the winner of this game given this board setup.\\
Your answer should specify the winner as one of the following strings: "White", "Black", or "Draw".\\
\\
PGN: 1. d4 d5 2. e3 e6 3. Bd3 Nf6 4. Nd2 Be7 5. c3 O-O 6. f4 Nbd7 7. Qe2 c5 8. Ngf3 c4 9. Bc2 a6 10. O-O b5 11. Ne5 Bb7 12. a3 Rb8 13. e4 dxe4 14. Nxe4 Nxe5 15. fxe5 Nd5 16. Qg4 a5 17. Bh6 f6 18. Qxg7\#\\
\\
\textbf{Instructions:}\\
- Provide your 3 best guesses and the probability that each is correct (0\% to 100\%) for the following question.\\
- If the number of options for the question is less than 3, only provide the confidence score for each option as the answer.\\
- Please reason step by step\\
- At the end, present your final answers and confidence scores in the following XML format:\\
<answer1>[only the winner of this game: "White", "Black", or "Draw"]</answer1>\\
<confidence1>[your confidence score for the answer1]</confidence1>\\
<answer2>[only the winner of this game: "White", "Black", or "Draw"]</answer2>\\
<confidence2>[your confidence score for the answer2]</confidence2>\\
<answer3>[only the winner of this game: "White", "Black", or "Draw"]</answer3>\\
<confidence3>[your confidence score for the answer3]</confidence3>\\
\\
\textbf{Example output:}\\
\char91 YOUR\_REASONING\char93\\
<answer1>Draw</answer1>\\
<confidence1>90\%</confidence1>\\
<answer2>Black</answer2>\\
<confidence2>80\%</confidence2>\\
<answer3>White</answer3>\\
<confidence3>50\%</confidence3>\\
\end{tcolorbox}

\begin{tcolorbox}[remarkbox, title=image\_math\_parity\_topk]
\small
You are given a plot of a real-valued, scalar function f(x).\\
YOUR TASK is to determine whether f(x) is an even function, an odd function, or neither.\\
- Definition of an even function: A function such that f(x) = f(-x) where the value remains unchanged if the sign of the independent variable is reversed.\\
- Definition of an odd function: A function such that f(-x) = -f(x) where the sign is reversed but the absolute value remains the same if the sign of the independent variable is reversed\\
- A function is neither even nor odd if it does not satisfy either definitions.\\
\\
\textbf{Instructions:}\\
- Provide your 3 best guesses and the probability that each is correct (0\% to 100\%) for the following question.\\
- If the number of options for the question is less than 3, only provide the confidence score for each option as the answer.\\
- Please reason step by step\\
- At the end, present your final answers and confidence scores in the following XML format:\\
<answer1>[only the final result: 'even', 'odd', or 'neither']</answer1>\\
<confidence1>[your confidence score for the answer1]</confidence1>\\
<answer2>[only the final result: 'even', 'odd', or 'neither']</answer2>\\
<confidence2>[your confidence score for the answer2]</confidence2>\\
<answer3>[only the final result: 'even', 'odd', or 'neither']</answer3>\\
<confidence3>[your confidence score for the answer3]</confidence3>\\
\\
\textbf{Example output:}\\
\char91 YOUR\_REASONING\char93\\
<answer1>even</answer1>\\
<confidence1>90\%</confidence1>\\
<answer2>odd</answer2>\\
<confidence2>80\%</confidence2>\\
<answer3>neither</answer3>\\
<confidence3>50\%</confidence3>\\
\end{tcolorbox}

\begin{tcolorbox}[remarkbox, title=image\_math\_convexity\_topk]
\small
You are given a plot of a real-valued, scalar function f(x).\\
YOUR TASK is to determine whether f(x) is an convex function or a concave function\\
- Definition of a convex function: A function such that for all x, y, and $0 \leq t \leq 1$\\
$f(tx + (1 - t)y) \leq tf(x) + (1 - t)f(y)$\\
- Definition of a concave function: A function such that for all x, y, and $0 \leq t \leq 1$\\
$f(tx + (1 - t)y) \geq tf(x) + (1 - t)f(y)$\\
\\
\textbf{Instructions:}\\
- Provide your 3 best guesses and the probability that each is correct (0\% to 100\%) for the following question.\\
- If the number of options for the question is less than 3, only provide the confidence score for each option as the answer.\\
- Please reason step by step\\
- At the end, present your final answers and confidence scores in the following XML format:\\
<answer1>[only the final result: 'convex' or 'concave']</answer1>\\
<confidence1>[your confidence score for the answer1]</confidence1>\\
<answer2>[only the final result: 'convex' or 'concave']</answer2>\\
<confidence2>[your confidence score for the answer2]</confidence2>\\
\\
\textbf{Example output:}\\
\char91 YOUR\_REASONING\char93\\
<answer1>convex</answer1>\\
<confidence1>80\%</confidence1>\\
<answer2>concave</answer2>\\
<confidence2>60\%</confidence2>\\
\end{tcolorbox}

\begin{tcolorbox}[remarkbox, title=image\_math\_breakpoint\_topk]
\small
You are given a plot of a real-valued, scalar function f(x).\\
YOUR TASK is to count the number of breakpoints in the plot of f(x). \space \\
A breakpoint refers to a point on the function’s domain at which the function changes its slope.\\
\\
You should IGNORE the left and right end point of the domain, i.e. if the function is defined on [a, b], you should only consider the domain (a, b).\\
\\
\textbf{Instructions:}\\
- Provide your 3 best guesses and the probability that each is correct (0\% to 100\%) for the following question.\\
- If the number of options for the question is less than 3, only provide the confidence score for each option as the answer.\\
- Please reason step by step\\
- At the end, present your final answers and confidence scores in the following XML format:\\
<answer1>[the number of breakpoints (in Arabic digits)]</answer1>\\
<confidence1>[your confidence score for the answer1]</confidence1>\\
<answer2>[the number of breakpoints (in Arabic digits)]</answer2>\\
<confidence2>[your confidence score for the answer2]</confidence2>\\
<answer3>[the number of breakpoints (in Arabic digits)]</answer3>\\
<confidence3>[your confidence score for the answer3]</confidence3>\\
\\
\textbf{Example output:}\\
\char91 YOUR\_REASONING\char93\\
<answer1>2</answer1>\\
<confidence1>80\%</confidence1>\\
<answer2>5</answer2>\\
<confidence2>60\%</confidence2>\\
<answer3>8</answer3>\\
<confidence3>50\%</confidence3>\\
\end{tcolorbox}

\begin{tcolorbox}[remarkbox, title=text\_math\_parity\_topk]
\small
You are given a real-valued, scalar function f(x).\\
YOUR TASK is to determine whether f(x) is an even function, an odd function, or neither.\\
- Definition of an even function: A function such that f(x) = f(-x) where the value remains unchanged if the sign of the independent variable is reversed.\\
- Definition of an odd function: A function such that f(-x) = -f(x) where the sign is reversed but the absolute value remains the same if the sign of the independent variable is reversed\\
- A function is neither even nor odd if it does not satisfy either definitions.\\
\\
Here is the expression of f(x){domain}:\\
\{text\}\\
\\
\textbf{Instructions:}\\
- Provide your 3 best guesses and the probability that each is correct (0\% to 100\%) for the following question.\\
- If the number of options for the question is less than 3, only provide the confidence score for each option as the answer.\\
- Please reason step by step\\
- At the end, present your final answers and confidence scores in the following XML format:\\
<answer1>[only the final result: 'even', 'odd', or 'neither']</answer1>\\
<confidence1>[your confidence score for the answer1]</confidence1>\\
<answer2>[only the final result: 'even', 'odd', or 'neither']</answer2>\\
<confidence2>[your confidence score for the answer2]</confidence2>\\
<answer3>[only the final result: 'even', 'odd', or 'neither']</answer3>\\
<confidence3>[your confidence score for the answer3]</confidence3>\\
\\
\textbf{Example output:}\\
\char91 YOUR\_REASONING\char93\\
<answer1>even</answer1>\\
<confidence1>90\%</confidence1>\\
<answer2>odd</answer2>\\
<confidence2>80\%</confidence2>\\
<answer3>neither</answer3>\\
<confidence3>50\%</confidence3>\\
\end{tcolorbox}

\begin{tcolorbox}[remarkbox, title=text\_math\_convexity\_topk]
\small
You are given a real-valued, scalar function f(x).\\
YOUR TASK is to determine whether f(x) is an convex function or a concave function\\
- Definition of a convex function: A function such that for all x, y, and $0 \leq t \leq 1$\\
$f(tx + (1 - t)y) \leq tf(x) + (1 - t)f(y)$\\
- Definition of a concave function: A function such that for all x, y, and $0 \leq t \leq 1$\\
$f(tx + (1 - t)y) \geq tf(x) + (1 - t)f(y)$\\
\\
Here is the expression of f(x){domain}:\\
\{text\}\\
\\
\textbf{Instructions:}\\
- Provide your 3 best guesses and the probability that each is correct (0\% to 100\%) for the following question.\\
- If the number of options for the question is less than 3, only provide the confidence score for each option as the answer.\\
- Please reason step by step\\
- At the end, present your final answers and confidence scores in the following XML format:\\
<answer1>[only the final result: 'convex' or 'concave']</answer1>\\
<confidence1>[your confidence score for the answer1]</confidence1>\\
<answer2>[only the final result: 'convex' or 'concave']</answer2>\\
<confidence2>[your confidence score for the answer2]</confidence2>\\
\\
\textbf{Example output:}\\
\char91 YOUR\_REASONING\char93\\
<answer1>convex</answer1>\\
<confidence1>80\%</confidence1>\\
<answer2>concave</answer2>\\
<confidence2>60\%</confidence2>\\
\end{tcolorbox}

\begin{tcolorbox}[remarkbox, title=text\_math\_breakpoint\_topk]
\small
You are given a real-valued, scalar function f(x).\\
YOUR TASK is to count the number of breakpoints in the plot of f(x). \space \\
A breakpoint refers to a point on the function’s domain at which the function changes its slope.\\
\\
Here is the expression of f(x){domain}:\\
\{text\}\\
\\
You should IGNORE the left and right end point of the domain, i.e. if the function is defined on [a, b], you should only consider the domain (a, b).\\
\\
\textbf{Instructions:}\\
- Provide your 3 best guesses and the probability that each is correct (0\% to 100\%) for the following question.\\
- If the number of options for the question is less than 3, only provide the confidence score for each option as the answer.\\
- Please reason step by step\\
- At the end, present your final answers and confidence scores in the following XML format:\\
<answer1>[the number of breakpoints (in Arabic digits)]</answer1>\\
<confidence1>[your confidence score for the answer1]</confidence1>\\
<answer2>[the number of breakpoints (in Arabic digits)]</answer2>\\
<confidence2>[your confidence score for the answer2]</confidence2>\\
<answer3>[the number of breakpoints (in Arabic digits)]</answer3>\\
<confidence3>[your confidence score for the answer3]</confidence3>\\
\\
\textbf{Example output:}\\
\char91 YOUR\_REASONING\char93\\
<answer1>2</answer1>\\
<confidence1>80\%</confidence1>\\
<answer2>5</answer2>\\
<confidence2>60\%</confidence2>\\
<answer3>8</answer3>\\
<confidence3>50\%</confidence3>\\
\end{tcolorbox}

\begin{tcolorbox}[remarkbox, title=chemistry\_image\_topk]
\small
You are given an image of a chemistry diagram.\\
YOUR TASK is to read the question and select the correct answer from the provided options.\\
\\
Which solution has a higher concentration of green particles?\\
A. Solution B\\
B. neither; their concentrations are the same\\
C. Solution A\\
\\
\textbf{Instructions:}\\
- Provide your 3 best guesses and the probability that each is correct (0\% to 100\%) for the following question.\\
- If the number of options for the question is less than 3, only provide the confidence score for each option as the answer.\\
- Carefully read and analyze the problem.\\
- Reason through the solution step by step, if helpful.\\
- At the end, present your final answers and confidence scores in the following XML format:\\
<answer1>[only the letter of the correct answer]</answer1>\\
<confidence1>[your confidence score here]</confidence1>\\
<answer2>[only the letter of the correct answer]</answer2>\\
<confidence2>[your confidence score here]</confidence2>\\
<answer3>[only the letter of the correct answer]</answer3>\\
<confidence3>[your confidence score here]</confidence3>\\
\\
\textbf{Example output:}\\
\char91 YOUR\_REASONING\char93\\
<answer1>A</answer1>\\
<confidence1>80\%</confidence1>\\
<answer2>C</answer2>\\
<confidence2>50\%</confidence2>\\
<answer3>B</answer3>\\
<confidence3>30\%</confidence3>\\
\end{tcolorbox}

\begin{tcolorbox}[remarkbox, title=chemistry\_text\_topk]
\small
You are given a multiple-choice chemistry question.\\
YOUR TASK is to read the question and select the correct answer from the provided options.\\
\\
In Solution A and Solution B, the green particles represent the solute. The volume of the solvent in two containers are equal. Solution A and Solution B have the same number of green particles.\\
\\
Which solution has a higher concentration of green particles?\\
A. Solution B\\
B. neither; their concentrations are the same\\
C. Solution A\\
\\
\textbf{Instructions:}\\
- Provide your 3 best guesses and the probability that each is correct (0\% to 100\%) for the following question.\\
- If the number of options for the question is less than 3, only provide the confidence score for each option as the answer.\\
- Carefully read and analyze the problem.\\
- Reason through the solution step by step, if helpful.\\
- At the end, present your final answers and confidence scores in the following XML format:\\
<answer1>[only the letter of the correct answer]</answer1>\\
<confidence1>[your confidence score here]</confidence1>\\
<answer2>[only the letter of the correct answer]</answer2>\\
<confidence2>[your confidence score here]</confidence2>\\
<answer3>[only the letter of the correct answer]</answer3>\\
<confidence3>[your confidence score here]</confidence3>\\
\\
\textbf{Example output:}\\
\char91 YOUR\_REASONING\char93\\
<answer1>A</answer1>\\
<confidence1>80\%</confidence1>\\
<answer2>C</answer2>\\
<confidence2>50\%</confidence2>\\
<answer3>B</answer3>\\
<confidence3>30\%</confidence3>\\
\end{tcolorbox}

\begin{tcolorbox}[remarkbox, title=physics\_image\_topk]
\small
You are given an image of a physics diagram.\\
YOUR TASK is to read the question and select the correct answer from the provided options.\\
\\
During this time, thermal energy was transferred from () to ().\\
A. the surroundings . . . each salmon\\
B. each salmon . . . the surroundings\\
\\
\textbf{Instructions:}\\
- Provide your 3 best guesses and the probability that each is correct (0\% to 100\%) for the following question.\\
- If the number of options for the question is less than 3, only provide the confidence score for each option as the answer.\\
- Carefully read and analyze the problem.\\
- Reason through the solution step by step, if helpful.\\
- At the end, present your final answer and a confidence score in the following XML format:\\
<answer1>[only the letter of the correct answer]</answer1>\\
<confidence1>[your confidence score here]</confidence1>\\
<answer2>[only the letter of the correct answer]</answer2>\\
<confidence2>[your confidence score here]</confidence2>\\
<answer3>[only the letter of the correct answer]</answer3>\\
<confidence3>[your confidence score here]</confidence3>\\
\\
\textbf{Example output:}\\
\char91 YOUR\_REASONING\char93\\
<answer1>A</answer1>\\
<confidence1>80\%</confidence1>\\
<answer2>C</answer2>\\
<confidence2>50\%</confidence2>\\
<answer3>B</answer3>\\
<confidence3>30\%</confidence3>\\
\end{tcolorbox}

\begin{tcolorbox}[remarkbox, title=physics\_text\_topk]
\small
You are given a multiple-choice physics question.\\
YOUR TASK is to read the question and select the correct answer from the provided options.\\
\\
The temperature of each salmon increased.\\
\\
During this time, thermal energy was transferred from () to ().\\
A. the surroundings . . . each salmon\\
B. each salmon . . . the surroundings\\
\\
\textbf{Instructions:}\\
- Provide your 3 best guesses and the probability that each is correct (0\% to 100\%) for the following question.\\
- If the number of options for the question is less than 3, only provide the confidence score for each option as the answer.\\
- Carefully read and analyze the problem.\\
- Reason through the solution step by step, if helpful.\\
- At the end, present your final answer and a confidence score in the following XML format:\\
<answer1>[only the letter of the correct answer]</answer1>\\
<confidence1>[your confidence score here]</confidence1>\\
<answer2>[only the letter of the correct answer]</answer2>\\
<confidence2>[your confidence score here]</confidence2>\\
<answer3>[only the letter of the correct answer]</answer3>\\
<confidence3>[your confidence score here]</confidence3>\\
\\
\textbf{Example output:}\\
\char91 YOUR\_REASONING\char93\\
<answer1>A</answer1>\\
<confidence1>80\%</confidence1>\\
<answer2>C</answer2>\\
<confidence2>50\%</confidence2>\\
<answer3>B</answer3>\\
<confidence3>30\%</confidence3>\\
\end{tcolorbox}

\section{Full Experimental Results}
In this section, we provide full experimental results as follows.
\subsection{MMMU-Pro}
The following table (\autoref{tab:appendix_mmmupro}) shows the full experimental results on MMLU-Pro.

\begin{table*}[!h]
\centering
\resizebox{\textwidth}{!}{
\begin{tabular}{llccc}
\toprule
Metric & Model & MMMU-Pro (Standard, 4) & MMMU-Pro (Standard, 10) & MMMU-Pro (Vision)  \\
\midrule
\multirow{13}{*}{ACC $\uparrow$}
       & o3                    & 80.4 & 73.7  & 67.6            \\
       & o4-mini               & 78.5 & 68.7  & 66.3            \\
       & o1                    & 77.7 & 70.4  & 64.4            \\
       & GPT4.1                & 73.4 & 65.0  & 59.5            \\
       & GPT4o                 & 67.8 & 57.7  & 53.8            \\
       & Qwen2-VL 7B           & 44.9 & 32.7  & 27.5            \\
       & Qwen2-VL 72B          & 59.0 & 48.5  & 42.1            \\
       & Qwen2.5-VL 7B         & 52.8 & 38.7  & 37.3            \\
       & Qwen2.5-VL 72B        & 64.4 & 53.8  & 49.7            \\
       & InternVL3 78B         & 65.5 & 55.1  & 44.5            \\
       & Skywork-R1V 38B       & 60.1 & 48.7  & 37.0            \\
       & Skywork-R1V2 38B      & 69.0 & 55.2  & 44.8            \\
       & Kimi-VL-A3B-Instruct  & 50.9 & 38.7  & 32.5            \\
       & Kimi-VL-A3B-Thinking  & 55.9    & 45.2  & 37.6            \\
\midrule
\multirow{13}{*}{ECE $\downarrow$}
       & o3                    & 0.060 & 0.047 & 0.077  \\
       & o4-mini               & 0.098 & 0.174 & 0.195  \\
       & o1                    & 0.174 & 0.245 & 0.303  \\
       & GPT4.1                & 0.231 & 0.321 & 0.373  \\
       & GPT4o                 & 0.279 & 0.383 & 0.420  \\
       & Qwen2-VL 7B           & 0.459 & 0.594 & 0.669  \\
       & Qwen2-VL 72B          & 0.293 & 0.411 & 0.476  \\
       & Qwen2.5-VL 7B         & 0.344 & 0.496 & 0.516  \\
       & Qwen2.5-VL 72B        & 0.280 & 0.392 & 0.431  \\
       & InternVL3 78B         & 0.280 & 0.387 & 0.496  \\
       & Skywork-R1V 38B       & 0.272 & 0.379 & 0.457  \\
       & Skywork-R1V2 38B      & 0.201 & 0.312 & 0.397  \\
       & Kimi-VL-A3B-Instruct  & 0.360 & 0.492 & 0.581  \\
       & Kimi-VL-A3B-Thinking  & 0.340     & 0.433 & 0.496  \\
\midrule
\multirow{13}{*}{AUROC $\uparrow$}
       & o3                    & 0.822 & 0.800 & 0.803  \\
       & o4-mini               & 0.755 & 0.790 & 0.763  \\
       & o1                    & 0.731 & 0.700 & 0.697  \\
       & GPT4.1                & 0.685 & 0.689 & 0.691  \\
       & GPT4o                 & 0.629 & 0.623 & 0.627  \\
       & Qwen2-VL 7B           & 0.505 & 0.534 & 0.523  \\
       & Qwen2-VL 72B          & 0.633 & 0.636 & 0.648  \\
       & Qwen2.5-VL 7B         & 0.599 & 0.593 & 0.593  \\
       & Qwen2.5-VL 72B        & 0.639 & 0.658 & 0.655  \\
       & InternVL3 78B         & 0.634 & 0.634 & 0.638  \\
       & Skywork-R1V 38B       & 0.645 & 0.671 & 0.669  \\
       & Skywork-R1V2 38B      & 0.683 & 0.705 & 0.680  \\
       & Kimi-VL-A3B-Instruct  & 0.534 & 0.560 & 0.598  \\
       & Kimi-VL-A3B-Thinking  & 0.589     & 0.638 & 0.660  \\
\bottomrule
\end{tabular}}%
\caption{MMMU-Pro results with CoT prompting. We evaluated the models following three settings: standard with 4 options, standard with 10 options, and vision-only with 10 options. All accuracy (Acc) values are in percentage.}
\label{tab:appendix_mmmupro}
\end{table*}

\subsection{Visual SimpleQA}
The following table (\autoref{tab:appendix_visualsimpleqa}) shows the full experimental results on Visual SimpleQA.
\begin{table*}[!h]
\centering
\resizebox{\textwidth}{!}{
\begin{tabular}{llcc}
\toprule
Metric & Model & Visual SimpleQA (text-only) & Visual SimpleQA (multimodal)  \\
\midrule
\multirow{12}{*}{ACC $\uparrow$}
       & o3                    & 80.5 & 73.6            \\
       & o4-mini               & 70.1 & 66.5            \\
       & o1                    & 80.0 & 70.6            \\
       & GPT4.1                & 80.2 & 67.1            \\
       & GPT4o                 & 76.8 & 63.5            \\
       & Qwen2-VL 7B           & 50.3 & 25.8            \\
       & Qwen2-VL 72B          & 59.0 & 45.4            \\
       & Qwen2.5-VL 7B         & 45.7 & 32.2            \\
       & Qwen2.5-VL 72B        & 61.6 & 49.6            \\
       & InternVL3 78B         & 57.3 & 44.0            \\
       & Kimi-VL-A3B-Instruct  & 50.5 & 35.3            \\
       & Kimi-VL-A3B-Thinking  & 47.5 & 39.3            \\
       & Skywork-R1V2-38B      & 60.3 & 45.5            \\
\midrule
\multirow{12}{*}{ECE $\downarrow$}
       & o3                    & 0.117 & 0.085  \\
       & o4-mini               & 0.112 & 0.069  \\
       & o1                    & 0.099 & 0.145            \\
       & GPT4.1                & 0.153 & 0.275            \\
       & GPT4o                 & 0.143 & 0.226            \\
       & Qwen2-VL 7B           & 0.454 & 0.603            \\
       & Qwen2-VL 72B          & 0.275 & 0.272            \\
       & Qwen2.5-VL 7B         & 0.381 & 0.252            \\
       & Qwen2.5-VL 72B        & 0.257 & 0.371            \\
       & InternVL3 78B         & 0.335 & 0.402            \\
       & Kimi-VL-A3B-Instruct  & 0.417 & 0.421            \\
       & Kimi-VL-A3B-Thinking  & 0.409 & 0.476            \\
       & Skywork-R1V2-38B      & 0.221 & 0.365            \\
\midrule
\multirow{12}{*}{AUROC $\uparrow$}
       & o3                    & 0.922 & 0.844  \\
       & o4-mini               & 0.903 & 0.859  \\
       & o1                    & 0.843 & 0.795            \\
       & GPT4.1                & 0.761 & 0.700            \\
       & GPT4o                 & 0.739 & 0.768            \\
       & Qwen2-VL 7B           & 0.540 & 0.567            \\
       & Qwen2-VL 72B          & 0.673 & 0.707            \\
       & Qwen2.5-VL 7B         & 0.712 & 0.785            \\
       & Qwen2.5-VL 72B        & 0.736 & 0.564            \\
       & InternVL3 78B         & 0.705 & 0.708            \\
       & Kimi-VL-A3B-Instruct  & 0.598 & 0.710            \\
       & Kimi-VL-A3B-Thinking  & 0.723 & 0.734            \\
       & Skywork-R1V2-38B      & 0.868 & 0.780            \\
\bottomrule
\end{tabular}}%
\caption{Visual SimpleQA results with CoT prompting. We evaluated the models in two settings: text-only and multimodal inputs. All accuracy (Acc) values are in percentage.}
\label{tab:appendix_visualsimpleqa}
\end{table*}

\subsection{Math}
The following table (\autoref{tab:appendix_math}) shows the full experimental results on MathVista and MathVision.
\begin{table*}[!h]
\centering
\resizebox{0.7\textwidth}{!}{
\begin{tabular}{llcc}
\toprule
Metric & Model & MathVista & MathVision \\
\midrule
\multirow{11}{*}{ACC $\uparrow$}
       & o3                          & 50.0       & 56.0        \\
       & o4-mini                     & 48.5       & 52.4        \\
       & o1                          & 46.7       & 51.0        \\
       & GPT4.1                      & 47.9       & 43.0        \\
       & GPT4o                       & 42.3       & 33.6        \\
       & Qwen2-VL 7B                 & 40.0       & 18.3        \\
       & Qwen2-VL 72B                & 45.0       & 28.7        \\
       & Qwen2.5-VL 7B               & 45.5       & 24.9        \\
       & Qwen2.5-VL 72B              & 49.9       & 40.8        \\
       & InternVL3 78B               & 47.1       & 34.2        \\
       & Skywork-R1V-38B             & 42.1       & 41.7        \\
       & Skywork-R1V2-38B            & 44.7       & 39.6        \\
       & Kimi-VL-A3B-Instruct        & 42.6       & 28.3        \\
       & Kimi-VL-A3B-Thinking        & 46.7       & 30.5        \\
\midrule
\multirow{11}{*}{ECE $\downarrow$}
       & o3                          & 0.242      & 0.111       \\
       & o4-mini                     & 0.388      & 0.327       \\
       & o1                          & 0.474      & 0.447       \\
       & GPT4.1                      & 0.485      & 0.535       \\
       & GPT4o                       & 0.518      & 0.618       \\
       & Qwen2-VL 7B                 & 0.477      & 0.793       \\
       & Qwen2-VL 72B                & 0.463      & 0.661       \\
       & Qwen2.5-VL 7B               & 0.418      & 0.660       \\
       & Qwen2.5-VL 72B              & 0.466      & 0.580       \\
       & InternVL3 78B               & 0.480      & 0.617       \\
       & Skywork-R1V-38B             & 0.470      & 0.455       \\
       & Skywork-R1V2-38B            & 0.438      & 0.427       \\
       & Kimi-VL-A3B-Instruct        & 0.482      & 0.612       \\
       & Kimi-VL-A3B-Thinking        & 0.475      & 0.598       \\
\midrule
\multirow{11}{*}{AUROC $\uparrow$}
       & o3                          & 0.586      & 0.822       \\
       & o4-mini                     & 0.575      & 0.724       \\
       & o1                          & 0.578      & 0.676       \\
       & GPT4.1                      & 0.632      & 0.674       \\
       & GPT4o                       & 0.611      & 0.599       \\
       & Qwen2-VL 7B                 & 0.564      & 0.551       \\
       & Qwen2-VL 72B                & 0.648      & 0.497       \\
       & Qwen2.5-VL 7B               & 0.611      & 0.547       \\
       & Qwen2.5-VL 72B              & 0.581      & 0.510       \\
       & InternVL3 78B               & 0.612      & 0.576       \\
       & Skywork-R1V-38B             & 0.605      & 0.574       \\
       & Skywork-R1V2-38B            & 0.619      & 0.757       \\
       & Kimi-VL-A3B-Instruct        & 0.583      & 0.549       \\
       & Kimi-VL-A3B-Thinking        & 0.625      & 0.603       \\
\bottomrule
\end{tabular}}%
\caption{MathVista and MathVision results with CoT prompting. All results are obtained on the ``testmini'' split of both datasets. All accuracy (Acc) values are in percentage.}
\label{tab:appendix_math}
\end{table*}

\subsection{Isobench}
The following table (\autoref{tab:appendix_isobench}) shows the full experimental results on IsoBench.
\begin{table*}[!h]
\centering
\resizebox{1.0\textwidth}{!}{
\begin{tabular}{ll*{5}{c}}
\toprule
Metric & Model & \multicolumn{5}{c}{IsoBench} \\
\cmidrule(lr){3-7}
& & Mathematics & Games & Science & Algorithms & All \\
\midrule
\multirow{10}{*}{Acc $\uparrow$}
& o3 & 89.5/99.3 & 52.5/63.2 & 93.3/98.0 & 84.5/98.3 & 79.5/89.7 \\
& o4-mini & 85.6/99.2 & 50.3/58.3 & 93.3/94.0 & 85.9/98.9 & 78.1/88.8 \\
& o1 & 76.0/99.2 & 40.7/51.0 & 91.3/96.7 & 76.6/98.2 & 69.0/87.2 \\
& GPT4.1 & 84.2/99.2 & 45.4/51.0 & 91.9/96.6 & 82.8/91.4 & 75.2/85.9 \\
& GPT4o & 73.2/98.0 & 38.8/46.4 & 85.2/98.7 & 64.9/72.9 & 64.2/80.7 \\
& Qwen2-VL 7B & 61.6/55.1 & 25.2/19.3 & 71.1/78.1 & 40.1/35.6 & 49.2/44.2 \\
& Qwen2-VL 72B & 67.1/94.0 & 26.9/34.5 & 82.7/93.2 & 49.3/47.8 & 55.0/70.4 \\
& Qwen2.5-VL 7B & 54.0/89.8 & 28.4/31.9 & 78.0/86.4 & 44.0/49.1 & 47.7/67.3 \\
& Qwen2.5-VL 72B & 60.3/99.1 & 24.1/41.1 & 86.0/95.3 & 61.2/66.8 & 53.7/78.2 \\
& InternVL3 78B & 61.4/98.8 & 24.5/38.2 & 88.7/96.0 & 58.9/64.0 & 54.1/76.8 \\
& Skywork-R1V-38B & 53.2/98.8 & 23.0/47.1 & 78.8/96.7 & 51.1/84.2 & 47.0/83.7 \\
& Skywork-R1V2-38B & 64.2/98.9 & 26.4/48.5 & 83.9/97.3 & 50.8/79.7 & 53.8/82.7 \\
& Kimi-VL-A3B-Instruct & 57.0/68.6 & 27.1/31.7 & 83.3/90.0 & 46.1/43.1 & 49.8/56.2 \\
& Kimi-VL-A3B-Thinking & 56.1/88.5 & 23.4/36.4 & 87.3/94.0 & 41.3/41.1 & 48.5/67.7 \\
\midrule
\multirow{10}{*}{ECE $\downarrow$}
& o3 & 0.088/0.075 & 0.162/0.106 & 0.094/0.113 & 0.034/0.085 & 0.037/0.081 \\
& o4-mini & 0.054/0.025 & 0.283/0.309 & 0.022/0.026 & 0.083/0.034 & 0.110/0.058 \\
& o1 & 0.187/0.007 & 0.522/0.425 & 0.080/0.033 & 0.223/0.022 & 0.265/0.109 \\
& GPT4.1 & 0.111/0.007 & 0.520/0.460 & 0.081/0.024 & 0.165/0.079 & 0.216/0.131 \\
& GPT4o & 0.230/0.018 & 0.593/0.499 & 0.146/0.011 & 0.335/0.260 & 0.332/0.178 \\
& Qwen2-VL 7B & 0.236/0.370 & 0.649/0.742 & 0.258/0.209 & 0.493/0.474 & 0.389/0.470 \\
& Qwen2-VL 72B & 0.292/0.055 & 0.652/0.549 & 0.158/0.031 & 0.501/0.494 & 0.408/0.258 \\
& Qwen2.5-VL 7B & 0.377/0.069 & 0.695/0.629 & 0.155/0.059 & 0.526/0.469 & 0.467/0.284 \\
& Qwen2.5-VL 72B & 0.353/0.008 & 0.724/0.534 & 0.146/0.039 & 0.376/0.320 & 0.431/0.200 \\
& InternVL3 78B & 0.339/0.007 & 0.684/0.539 & 0.110/0.042 & 0.386/0.307 & 0.412/0.198 \\
& Skywork-R1V-38B & 0.337/0.013 & 0.593/0.438 & 0.149/0.046 & 0.432/0.119 & 0.407/0.120 \\
& Skywork-R1V2-38B & 0.304/0.008 & 0.529/0.368 & 0.094/0.025 & 0.420/0.151 & 0.364/0.120 \\
& Kimi-VL-A3B-Instruct & 0.327/0.257 & 0.651/0.604 & 0.068/0.032 & 0.471/0.503 & 0.411/0.372 \\
& Kimi-VL-A3B-Thinking & 0.376/0.077 & 0.708/0.611 & 0.117/0.044 & 0.554/0.540 & 0.462/0.284 \\
\midrule
\multirow{10}{*}{AUROC $\uparrow$}
& o3 & 0.551/0.391 & 0.802/0.857 & 0.884/0.763 & 0.719/0.878 & 0.751/0.933 \\
& o4-mini & 0.500/0.254 & 0.847/0.835 & 0.705/0.772 & 0.790/0.662 & 0.737/0.840 \\
& o1 & 0.509/0.495 & 0.744/0.814 & 0.827/0.494 & 0.626/0.518 & 0.669/0.884 \\
& GPT4.1 & 0.610/0.470 & 0.694/0.745 & 0.538/0.547 & 0.593/0.668 & 0.569/0.773 \\
& GPT4o & 0.473/0.465 & 0.680/0.728 & 0.528/0.331 & 0.591/0.627 & 0.533/0.743 \\
& Qwen2-VL 7B & 0.541/0.579 & 0.541/0.468 & 0.522/0.555 & 0.457/0.383 & 0.495/0.555 \\
& Qwen2-VL 72B & 0.455/0.557 & 0.570/0.773 & 0.543/0.730 & 0.511/0.558 & 0.542/0.706 \\
& Qwen2.5-VL 7B & 0.464/0.489 & 0.620/0.666 & 0.552/0.565 & 0.576/0.593 & 0.470/0.586 \\
& Qwen2.5-VL 72B & 0.461/0.495 & 0.662/0.715 & 0.500/0.603 & 0.570/0.541 & 0.550/0.734 \\
& InternVL3 78B & 0.400/0.421 & 0.535/0.771 & 0.548/0.515 & 0.565/0.625 & 0.546/0.797 \\
& Skywork-R1V-38B & 0.501/0.431 & 0.588/0.736 & 0.654/0.528 & 0.531/0.726 & 0.584/0.781 \\
& Skywork-R1V2-38B & 0.525/0.566 & 0.608/0.851 & 0.592/0.591 & 0.583/0.766 & 0.617/0.885 \\
& Kimi-VL-A3B-Instruct & 0.508/0.430 & 0.617/0.724 & 0.629/0.715 & 0.582/0.661 & 0.523/0.597 \\
& Kimi-VL-A3B-Thinking & 0.482/0.632 & 0.586/0.487 & 0.465/0.522 & 0.517/0.585 & 0.509/0.507 \\
\bottomrule
\end{tabular}}
\caption{IsoBench results using CoT prompting. Image/text modality results are shown with slash (/). For Mathematics, the text modality shows LaTeX format results; for Games, the text modality shows PGN format results. All accuracy (Acc) values are in percentage.}
\label{tab:appendix_isobench}
\end{table*}

\section{Video}
The following table (\autoref{tab:appendix_videommmu}) shows the full experimental results on VideoMMMU.
\begin{table*}[!h]
\centering
\begin{tabular}{llc} 
\toprule
Metric & Model & VideoMMMU \\
\midrule
\multirow{13}{*}{ACC $\uparrow$}
       & o3              & 74.6 / 71.2 / 41.7 \\
       & o4-mini        & 72.8 / 67.5 / 39.2 \\
       & o1              & 72.7 / 66.2 / 40.3 \\
       & GPT4.1          & 74.9 / 62.3 / 40.8 \\
       & GPT4o           & 65.3 / 60.5 / 38.4 \\
       & LLaVA-OV 72B                & 44.4 / 33.3 / 28.8 \\
       & Qwen2.5-VL 7B               & 66.9 / 52.3 / 31.2 \\
       & Qwen2.5-VL 72B              & 80.0 / 69.7 / 44.3 \\
       & InternVL3 78B               & 66.7 / 54.7 / 35.7 \\
       & Kimi-VL-A3B-Instruct        & 72.3 / 41.7 / 30.7 \\
       & Kimi-VL-A3B-Thinking        & 62.0 / 55.1 / 32.6 \\
       & Skywork-R1V2-38B            & 59.9 / 58.6 / 40.7 \\
\midrule
\multirow{13}{*}{ECE $\downarrow$}
       & o3             & 0.073 / 0.051 / 0.092 \\
       & o4-mini         & 0.125 / 0.172 / 0.293 \\
       & o1               & 0.204 / 0.271 / 0.470 \\
       & GPT4.1          & 0.225 / 0.342 / 0.549 \\
       & GPT4o           & 0.311 / 0.349 / 0.519 \\
       & LLaVA-OV 72B                & 0.493 / 0.570 / 0.618 \\
       & Qwen2.5-VL 7B               & 0.235 / 0.347 / 0.549 \\
       & Qwen2.5-VL 72B              & 0.161 / 0.248 / 0.443 \\
       & InternVL3 78B               & 0.278 / 0.379 / 0.551 \\
       & Kimi-VL-A3B-Instruct        & 0.203 / 0.488 / 0.571 \\
       & Kimi-VL-A3B-Thinking        & 0.343 / 0.371 / 0.553 \\
       & Skywork-R1V2-38B            & 0.233 / 0.271 / 0.403 \\
\midrule
\multirow{13}{*}{AUROC $\uparrow$}
       & o3              & 0.794 / 0.752 / 0.796 \\
       & o4-mini         & 0.745 / 0.697 / 0.738 \\
       & o1              & 0.659 / 0.657 / 0.723 \\
       & GPT4.1          & 0.648 / 0.726 / 0.568 \\
       & GPT4o           & 0.587 / 0.582 / 0.642 \\
       & LLaVA-OV 72B                & 0.527 / 0.522 / 0.461 \\
       & Qwen2.5-VL 7B               & 0.590 / 0.582 / 0.610 \\
       & Qwen2.5-VL 72B              & 0.645 / 0.701 / 0.626 \\
       & InternVL3 78B               & 0.578 / 0.613 / 0.529 \\
       & Kimi-VL-A3B-Instruct        & 0.621 / 0.626 / 0.531 \\
       & Kimi-VL-A3B-Thinking        & 0.577 / 0.625 / 0.555 \\
       & Skywork-R1V2-38B            & 0.740 / 0.675 / 0.666 \\
\bottomrule
\end{tabular}%
\caption{VideoMMMU results using CoT prompting. All results were tested by using 32 frames uniformly sampled from the video. The scores are reported in the order of Perception/Comprehension/Adaptation splits. All accuracy (Acc) values are in percentage.} 
\label{tab:appendix_videommmu}
\end{table*}

\newpage
\section{Diffferent Prompting Strategies}
The following tables (\autoref{tab:appendix_top3_prompt} and \autoref{appendix:self_reflection}) show the performances with different prompting strategies on IsoBench.
\begin{table*}[htbp] 
\centering 
\begin{tabular}{ll*{5}{c}} 
\toprule 
Metric & Model & \multicolumn{5}{c}{IsoBench} \\ 
\cmidrule(lr){3-7} 
& & Mathematics & Games & Science & Algorithms & All \\ 
\midrule 
\multirow{6}{*}{Acc $\uparrow$} 
& GPT-4.1 & 84.5/99.0 & 44.3/51.1 & 91.9/97.3 & 82.0/91.6 & 75.0/86.0 \\
& Qwen2-VL 7B & 56.9/53.8 & 18.8/24.7 & 65.7/74.8 & 46.7/37.5 & 46.4/44.3 \\ 
& Qwen2-VL 72B & 66.3/89.8 & 26.0/35.5 & 82.0/92.7 & 45.5/50.7 & 53.7/69.6 \\ 
& Qwen2.5-VL 7B & 57.5/87.1 & 28.1/34.2 & 73.3/81.1 & 47.1/49.3 & 49.6/66.4 \\ 
& Qwen2.5-VL 72B & 56.5/98.9 & 24.3/42.7 & 86.0/96.7 & 56.8/72.4 & 51.1/79.7 \\ 
& InternVL3 78B & 62.7/98.4 & 24.5/40.9 & 91.3/94.7 & 58.6/64.8 & 54.9/77.4 \\ 
\midrule 
\multirow{6}{*}{ECE $\downarrow$} 
& GPT-4.1 & 0.092/0.013 & 0.502/0.426 & 0.053/0.021 & 0.157/0.065 & 0.199/0.106 \\
& Qwen2-VL 7B & 0.364/0.446 & 0.760/0.716 & 0.343/0.235 & 0.530/0.610 & 0.490/0.536 \\ 
& Qwen2-VL 72B & 0.306/0.093 & 0.678/0.567 & 0.156/0.029 & 0.532/0.452 & 0.428/0.270 \\ 
& Qwen2.5-VL 7B & 0.365/0.122 & 0.702/0.616 & 0.231/0.170 & 0.503/0.495 & 0.462/0.319 \\ 
& Qwen2.5-VL 72B & 0.380/0.010 & 0.711/0.505 & 0.130/0.054 & 0.415/0.258 & 0.447/0.180 \\ 
& InternVL3 78B & 0.306/0.014 & 0.604/0.466 & 0.087/0.029 & 0.351/0.295 & 0.369/0.174 \\ 
\midrule 
\multirow{6}{*}{AUROC $\uparrow$} 
& GPT-4.1 & 0.541/0.249 & 0.708/0.796 & 0.554/0.720 & 0.680/0.534 & 0.580/0.751 \\
& Qwen2-VL 7B & 0.365/0.506 & 0.759/0.280 & 0.489/0.540 & 0.487/0.642 & 0.465/0.547 \\ 
& Qwen2-VL 72B & 0.464/0.545 & 0.590/0.794 & 0.546/0.707 & 0.549/0.531 & 0.555/0.713 \\ 
& Qwen2.5-VL 7B & 0.441/0.512 & 0.599/0.704 & 0.590/0.630 & 0.490/0.480 & 0.457/0.613 \\ 
& Qwen2.5-VL 72B & 0.494/0.591 & 0.605/0.721 & 0.472/0.345 & 0.533/0.583 & 0.537/0.785 \\ 
& InternVL3 78B & 0.397/0.526 & 0.474/0.727 & 0.584/0.706 & 0.607/0.661 & 0.579/0.822 \\ 
\bottomrule 
\end{tabular} 
\caption{IsoBench results using Top-K prompting, where K=3. Image/text modality results are shown with slash (/). For Mathematics, the text modality shows LaTeX format results; for Games, the text modality shows PGN format results. All accuracy (Acc) values are in percentage.} 
\label{tab:appendix_top3_prompt} 
\end{table*}

\begin{table*}[h] 
\centering 
\begin{tabular}{ll*{5}{c}} 
\toprule 
Metric & Model & \multicolumn{5}{c}{IsoBench} \\ 
\cmidrule(lr){3-7} 
& & Mathematics & Games & Science & Algorithms & All \\ 
\midrule 
\multirow{3}{*}{Acc $\uparrow$} 
& o3 & 90.0/99.2 & 54.5 / 64.1 & 93.2 / 97.3 & 84.9 / 100.0 & 80.7 / 90.7 \\
& o4-mini & 85.0 / 99.2 & 48.8 / 58.8 & 92.6 / 96.0 & 83.3 / 98.4 & 77.0 / 89.3 \\
& o1 & 75.0 / 99.2 & 37.0 / 47.7 & 91.8 / 97.3 & 70.3 / 97.9 & 67.3 / 87.3 \\
\midrule 
\multirow{3}{*}{ECE $\downarrow$} 
& o3 & 0.110 / 0.077 & 0.133 / 0.085 & 0.077 / 0.069 & 0.047 / 0.085 & 0.046 / 0.075 \\
& o4-mini & 0.077 / 0.020 & 0.130 / 0.174 & 0.057 / 0.007 & 0.089 / 0.017 & 0.089 / 0.031 \\
& o1 & 0.195 / 0.006 & 0.490 / 0.263 & 0.060 / 0.027 & 0.236 / 0.020 & 0.252 / 0.066 \\
\midrule 
\multirow{3}{*}{AUROC $\uparrow$} 
& o3 & 0.696 / 0.529 & 0.745 / 0.835 & 0.795 / 0.878 & 0.798 / - (all correct) & 0.772 / 0.942 \\
& o4-mini & 0.602 / 0.685 & 0.820 / 0.803 & 0.820 / 0.829 & 0.788 / 0.964 & 0.781 / 0.913 \\
& o1 & 0.583 / 0.633 & 0.829 / 0.855 & 0.712 / 0.580 & 0.715 / 0.571 & 0.740 / 0.939 \\
\bottomrule 
\end{tabular} 
\caption{IsoBench results using self-reflection prompting. Image/text modality results are shown with slash (/). For Mathematics, the text modality shows LaTeX format results; for Games, the text modality shows PGN format results. All accuracy (Acc) values are in percentage.} 
\label{appendix:self_reflection}
\end{table*}

\section{Generative AI Statement}
Large language models were utilized to facilitate aspects of the writing process in this project. Specifically, Claude Sonnet 3.7 was employed to assist with formatting components of the manuscript and generating templates for LaTeX tables and figures. All machine-generated content underwent thorough review, editing, and verification by the authors to maintain factual precision and academic integrity.

\end{document}